\documentclass[lettersize,journal]{IEEEtran}
\usepackage{amsmath,amsfonts}
\usepackage{algorithmic}
\usepackage{algorithm}
\usepackage{array}
\usepackage[caption=false,font=normalsize,labelfont=sf,textfont=sf]{subfig}
\usepackage{textcomp}
\usepackage{stfloats}
\usepackage{url}
\usepackage{verbatim}
\usepackage{graphicx}
\usepackage{cite}
\usepackage[utf8]{inputenc} 
\usepackage[T1]{fontenc}    
\usepackage{hyperref}       
\usepackage{url}            
\usepackage{booktabs}       
\usepackage{amsfonts}       
\usepackage{multirow}       
\usepackage{graphicx}
\usepackage{nicefrac}       
\usepackage{microtype}      
\usepackage{xcolor}         
\usepackage{enumitem}
\usepackage{adjustbox}
\usepackage[table]{xcolor}
\definecolor{lightgreen}{RGB}{220,245,220}
\newcommand{\srcell}[1]{\cellcolor{lightgreen}#1}

\begin{document}

\title{Beyond Visual Fidelity: Benchmarking Super-Resolution Models for Large-Scale Remote Sensing Imagery via Downstream Task Integration}

\author{
Zhili Li, Kangyang Chai, Zhihao Wang, Xiaowei Jia, Yanhua Li, Gengchen Mai, Sergii Skakun, \\
Zhe Zhu, Dinesh Manocha, Yiqun Xie$^*$

\thanks{
$^*$Corresponding author: Yiqun Xie.
}
\thanks{Zhili Li, Kangyang Chai, Zhihao Wang, Sergii Skakun, Dinesh Manocha, and Yiqun Xie are with University of Maryland.
E-mail:  \{lizhili, kychai,  zhwang1, skakun, dmanocha, xie\}@umd.edu.
Xiaowei Jia is with Rutgers University.
E-mail:  xj159@cs.rutgers.edu.
Yanhua Li is with Worcester Polytechnic Institute.
E-mail:  yli15@wpi.edu.
Gengchen Mai is with University of Texas at Austin.
E-mail:  gengchen.mai@austin.utexas.edu.
Zhe Zhu is with University of Connecticut. 
E-mail: zhe@uconn.edu}
\thanks{Manuscript under review at Remote Sensing of Environment (RSE).}
}

\markboth{Journal of \LaTeX\ Class Files,~Vol.~14, No.~8, August~2021}%
{Shell \MakeLowercase{\textit{et al.}}: A Sample Article Using IEEEtran.cls for IEEE Journals}

\maketitle

\begin{abstract}
Super-resolution (SR) techniques have made major advances in reconstructing high-resolution images from low-resolution inputs. The increased resolution provides visual enhancement and utility for monitoring tasks. In particular, SR has been increasingly developed for satellite-based Earth observation, with applications in urban planning, agriculture, ecology, and disaster response. However, existing SR studies and benchmarks typically use fidelity metrics such as PSNR or SSIM, whereas the true utility of super-resolved images lies in supporting downstream tasks such as land cover classification, biomass estimation, and change detection. To bridge this gap, we introduce GeoSR-Bench, a downstream task-integrated SR benchmark dataset to evaluate SR models beyond fidelity metrics. GeoSR-Bench comprises spatially co-located, temporally aligned, and quality-controlled image pairs from about 36,000 locations across diverse land covers, spanning resolutions from 500m to 0.6m. To the best of our knowledge, GeoSR-Bench is the first SR benchmark that directly connects improved image resolution from SR models with downstream Earth monitoring tasks, including land cover segmentation, infrastructure mapping, and biophysical variable estimation. Using GeoSR-Bench, we benchmark GAN, transformer, neural operator, and diffusion-based SR models on perceptual quality and downstream task performance. We conduct experiments with 270 settings, covering 2 cross-platform SR tasks, 9 SR models, 3 downstream task models, and 5 downstream tasks for each SR task. The results show that improvements in traditional SR metrics often do not correlate with gains in task performance, and the correlations can be negative, indicating that these metrics provide limited guidance for selecting superior models for downstream tasks. This reveals the need to integrate downstream tasks into SR model development and evaluation.
\end{abstract}


\section{Introduction}\label{sec:intro}

Resolution enhancement for remote sensing images brings promising potential to unlock the collective monitoring power offered by diverse platforms.
First, high-resolution remote sensing images (e.g., meter-level or sub-meter resolutions) play a vital role in a wide array of Earth monitoring tasks, 
including urban planning, agriculture, ecology, disaster response, and so on, where fine-granularity visual signals are important to capture details such as narrow roads \cite{zhao2024narrow}, individual residential buildings \cite{liu2023china,chen2023large}, vehicles, solar panels \cite{clark2023solar}, and fragmented land use. 
However, acquiring such imagery at large scale and continuously over time is often challenging due to the very high cost. In addition, such platforms are often provided by commercial sectors, making the data highly expensive and less accessible.
Second, many high-resolution platforms (e.g., PlanetScope) rely on more recent sensing technologies, such as small-satellite constellations with improved high-resolution optical sensors, which are only available in nearer time frame, significantly limiting data availability over time for tracking past changes and events \cite{fu2024remote}. In contrast, coarse (e.g., 200 to 1000m) and medium  (e.g., 10 to 30m) resolution platforms have much longer availability spanning over multiple past decades.
Lastly, in addition to high-resolution images, medium-resolution images are also important for a broad range of applications, such as crop mapping, yield prediction, deforestation monitoring, ice tracking, etc.
However, these sensing platforms often have a small constellation with a few satellites, reducing their revisit time to each location and increasing the temporal gaps. For example, the revisit time is about 16 days for Landsat-8 and 5-10 days for Sentinel-2 \cite{li2017global}.
On the other hand, coarse-resolution platforms have a much larger footprint per image and higher revisit frequency. For example, the MODIS platform can offer a daily global coverage.
Thus, upgrading the resolution both from coarse to medium and from medium to high
enables new opportunities to significantly improve the monitoring capabilities.

Super-resolution (SR) techniques have emerged as powerful tools to reconstruct high-resolution images from lower-resolution inputs.
However, existing developments of SR methods for remote sensing data have mostly focused on the visual 
fidelity 
of the generated images \cite{he2021spatial} (e.g., structural similarity to high-resolution images)
but lack emphasis on the downstream 
tasks, such as land cover classification, biomass estimation, and change detection \cite{kowaleczko2023real}.
In Earth observation, though, the visual fidelity and sharpness may not directly link to improvements in these related applications \cite{he2022tracking}, and spurious details--if present--may even lead to incorrect maps and negatively impact decision making.
Thus, this paper aims to bridge the gap by developing a new benchmark dataset integrating downstream tasks (e.g., land cover mapping, infrastructure masking, environment monitoring) into the evaluation of SR methods, providing a foundation for task-integrated SR development and enabling assessment of whether the sharper images can truly improve the solution quality on various downstream tasks.

There have been broad efforts in developing benchmark datasets for SR tasks, including several efforts on the remote sensing imagery side.
Benchmarks such as DIV2K \cite{agustsson2017ntire}, Flickr2K \cite{lim2017enhanced}, and LSDIR \cite{li2023lsdir} were developed for non-satellite images and 
support training using synthetic pairs, where low-resolution inputs are synthetically downsampled from high-resolution images.
Recent efforts including \cite{kohler2019toward} and \cite{cai2019toward}, have also improved that using real pairs of low and high resolution images which can be captured by, for example, changing the focus of the camera. 
However, these datasets focus on regular images that often have a very different view and resolution than remote sensing images, such as horizontal vs. top-down views, and millimeter vs meter level resolutions.
Expanding this, various efforts have developed benchmark datasets for remote sensing images, including SEN2VENµS \cite{michel2022sen2venmus}, WorldStrat \cite{cornebise2022open}, and PROBA-V \cite{martens2019super}.
However, these datasets considered only traditional metrics related to perception-level visual appearance, such as Peak Signal-to-Noise Ratio (PSNR) and Structural Similarity Index (SSIM), which do not reveal if and how the SR results can benefit tasks in downstream applications.

\begin{figure}[t]
  \centering
  \includegraphics[width=0.49\textwidth]{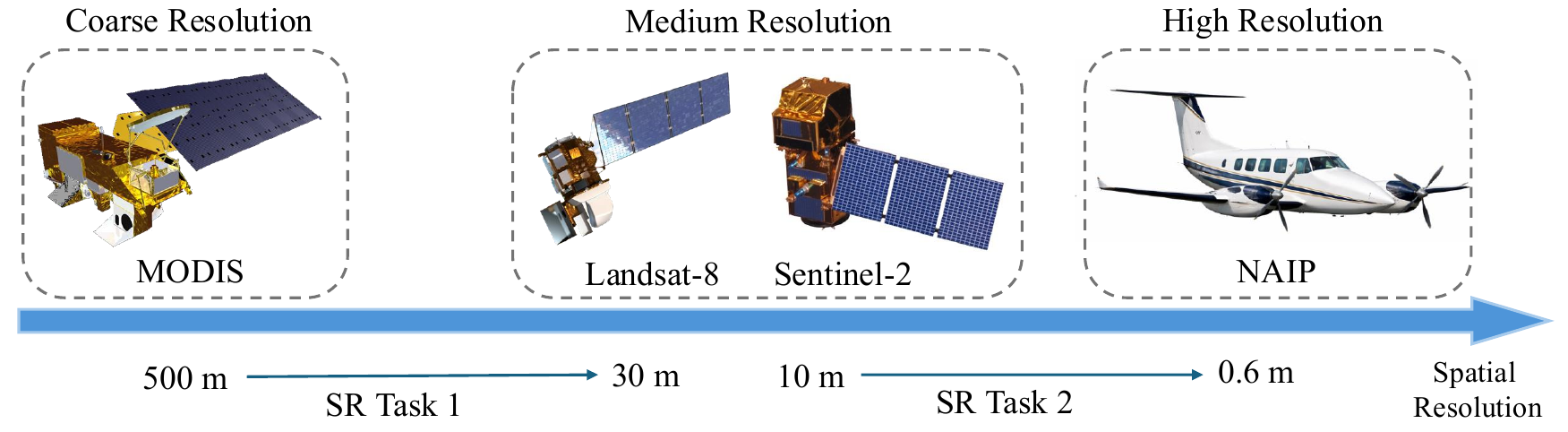}
  \caption{GeoSR benchmark focuses on two cross-platform SR tasks: MODIS (500m) to Landsat-8 (30m), and Sentinel-2 (10m) to NAIP (0.6m), covering platforms and resolutions that span the most widely used in downstream applications.}
  \label{fig:sr_settings}
\end{figure}

To bridge the gaps, we introduce GeoSR-Bench, a downstream task-integrated SR benchmark dataset to more deeply evaluate the performance of SR models beyond fidelity metrics. Our contributions are summarized as follows:
\begin{itemize}[leftmargin=*]
\item We develop \textbf{GeoSR-Bench}, the first benchmark dataset for remote sensing image super-resolution that integrates downstream tasks for direct evaluations in the application context.
The SR part of the dataset comprises spatially co-located, temporally aligned, and quality-controlled image pairs from approximately 20{,}000 locations over the global across diverse urban and non-urban land covers, which can be used to train and evaluate SR models using standard fidelity metrics.
The dataset covers multiple remote sensing platforms with a broad range of commonly used resolutions, from coarse to medium, and from medium to high. 
Specifically, it includes two cross-platform SR tasks (Fig. \ref{fig:sr_settings}): from MODIS to Landsat-8 (i.e., from 500m to 30m) and from Sentinel-2 to NAIP (i.e., from 10m to 0.6m), which collectively represent platforms that are most widely used in downstream applications.
\item GeoSR-Bench further includes 5 downstream tasks from different applications (e.g., land cover segmentation, infrastructure mapping, biophysical variable estimation) for each cross-platform SR task, such that higher-resolution imagery provides a clear advantage over lower-resolution inputs, enabling meaningful evaluation of SR improvements. The downstream task part contains image pairs from lower- and higher-resolution platforms as well as target pixel-level labels from approximately 16,000 locations across two cross-platform SR tasks (i.e., MODIS to Landsat-8 and Sentinel-2 to NAIP). 
These allow finetuning of SR models and training task-focused models to evaluate the real utility of SR results in the application context. 
\item We benchmark a suite of SR models, including GAN-, transformer-, neural operator-, and diffusion-based models, using metrics for both perceptual quality and downstream tasks. For perceptual quality, we consider two types of fidelity metrics: (1) traditional visual fidelity metrics, which assess pixel-level accuracy and local image similarity, including PSNR and SSIM; and (2) geostatistical spatial fidelity metrics based on semivariogram for spatial dependency \cite{qiu2023evaluation, song2025tif}. For downstream tasks, we evaluate the utility of SR images using task-specific performance metrics, including F1 score for classification tasks and mean absolute error (MAE) for regression tasks.
We conduct extensive experiments with 270 different settings, covering 2 cross-platform SR tasks, 9 SR models, 3 downstream task models, 5 downstream tasks for each SR task. The results reveal that improvements in visual and spatial fidelity SR metrics often do not correlate with gains in task performance, and the correlations can be negative in many scenarios, indicating that these metrics provide only limited guidance when concluding superior models for downstream tasks. This reveals the necessity and importance of integrating downstream
tasks into the continued development and evaluation of SR models.
\item We open-source the dataset, trained models, and evaluation protocols, to support the development of new SR models and evaluations with a variety of downstream tasks. The dataset, code, and trained models for this benchmark are publicly available at \url{https://github.com/ai-spatial/GeoSR-Bench}.
\end{itemize} 

\textit{\underline{Scope:}} This paper focuses on \textbf{pixel-level tasks} (e.g., semantic segmentation, pixel-level regression) for downstream applications because: (1) Pixel-level tasks are most common in real-world applications for Earth observation, where the fine-grained results are necessary for end-users to perform area estimation (e.g., acreage estimation for different types of crops in agriculture), property boundary mapping, land cover mapping, biomass estimation, and many more. For example,
the USDA Cropland Data Layer (CDL) \cite{usda_nass_cdl} provides per-pixel crop type classification
to support applications such as crop rotation pattern monitoring, agricultural expansion assessment, and crop yield prediction. The ESA WorldCover dataset \cite{esa_worldcover} provides global, per-pixel land cover mapping, enabling large-scale analyses of land use and land cover patterns, ecosystem distribution, and environmental change across different regions and years. \textbf{Here pixel-level means that results are needed for each pixel in the input image, but does not require individual pixels as inputs.} For example, semantic segmentation with U-Net falls into the pixel-level tasks here.
(2) Pixel-level tasks rely most on resolution and thus can best reflect the changes by SR models, which is important for evaluating the improvements to the downstream tasks.

\section{Related Work}

\subsection{Image Super-Resolution}
Traditional SR methods primarily follow task-agnostic paradigms, focusing on accurate reconstruction of high-resolution images from low-resolution inputs regardless of their usability on downstream applications. 
Early approaches include interpolation-based methods \cite{zhou2012interpolation, zhang2018single} and early learning-based techniques \cite{kim2010single}, followed by deep convolutional neural networks (CNN) such as SRCNN \cite{dong2014learning}, VDSR \cite{kim2016accurate}, EDSR \cite{lim2017enhanced}, and RCAN \cite{zhang2018image}. More recent work leverages transformer-based architectures such as TTSR \cite{yang2020learning}, SwinIR \cite{liang2021swinir}, ESRT \cite{lu2022transformer}, HAT \cite{chen2023activating}, and CFAT \cite{ray2024cfat}, which model long-range dependencies for improved reconstruction.
In parallel, generative adversarial network (GAN)-based models, such as SRGAN \cite{ledig2017photo}, ESRGAN \cite{wang2018esrgan}, Real-ESRGAN \cite{wang2021real}, and SeD \cite{li2024sed} focus on generating photorealistic textures during SR. Diffusion-based methods, such as SR3 \cite{saharia2022image}, DiffusionSR \cite{ho2022cascaded}, and DiffusionSat \cite{khanna2024diffusionsat}, introduce a stochastic iterative denoising process to produce sharper details. Most recently, neural operator-based models including SRNO \cite{wei2023super} and DiffFNO \cite{liu2025difffno} enable super-resolution at arbitrary scaling factors by modeling low‑ and high‑resolution images as continuous functions. These methods are commonly evaluated using metrics such as PSNR, SSIM \cite{wang2004image}, and LPIPS \cite{zhang2018unreasonable}, which assess pixel-level accuracy or perceptual similarity but do not necessarily reflect performance in downstream tasks. Although these approaches have achieved impressive visual results on benchmarks like DIV2K \cite{agustsson2017ntire} and Flickr2K \cite{lim2017enhanced}, their effectiveness in downstream tasks remains underexplored.

\subsection{Remote Sensing Image Super-Resolution}
Traditionally, the SR task for images from a single remote sensing platform is handled by an additional panchromatic band (pan-band), which is by design collected at a higher resolution to help improve the resolution of the other spectral bands. The related methods are also known as pan-sharpening.
Both rule-based (e.g., GSA \cite{aiazzi2007improving}, SFIM \cite{liu2000smoothing}) and ML methods (e.g., PanFormer \cite{zhou2022panformer}, MutInf \cite{zhou2022mutual}) pan-sharpening methods have been developed.
However, pan-sharpening is limited by the resolution of the pan-band, and the characteristics of pan-band can be different than other spectral bands. 
For example, the 30m-resolution images from Landsat-8 come with a pan-band at 15m, which is still in the medium resolution range and lacks spatial details.
To address this, image fusion methods have also been developed, which integrate images from satellite platforms with higher resolution into images captured by lower resolution platforms, such as DCSTFN \cite{tan2018deriving}, BiaSTF \cite{li2020new}, VIPSTF \cite{wang2020virtual}, and TIF \cite{song2025tif}. Specifically, TIF leverages long-term Landsat and Sentinel-2 time series to learn pixel-level cross-sensor spectral relationships and generate 10-m surface reflectance time series, whose utility is evaluated on a downstream multi-date change detection task \cite{song2025tif}.
Beyond traditional pan-sharpening and image fusion methods, many SR architectures originally developed for natural images have been adapted to the remote sensing domain. These include CNN-based models \cite{arun2020cnn, vasilescu2023cnn}, transformer-based models \cite{kang2024efficient, lei2021transformer}, diffusion-based models \cite{xiao2023ediffsr, meng2024conditional}, and neural operator-based models \cite{zhao2025arbitrary, xu2025neurop}.
More recently, foundation models have been developed for remote sensing images and several have considered SR tasks. For example, DiffusionSat \cite{khanna2024diffusionsat} generates high-resolution images based on a sequence of lower-resolution images with associated metadata and the caption that describes the target high-resolution image;  Text2Earth \cite{liu2025text2earth} guides the generation of super-resolved images by augmenting the foundation model with a trainable ControlNet \cite{zhang2023adding} module that encodes the low-resolution image as conditional guidance. 
However, these methods mainly focus on visual fidelity, where sharper boundaries of geo-objects or events can be generated in the higher-resolution results without connecting to downstream tasks, and the evaluation still relies on traditional metrics such as PSNR and SSIM.

\subsection{Benchmark Datasets for Super Resolution.}
Progress in SR research has been fueled by the availability of high-quality datasets.
Datasets such as DIV2K \cite{agustsson2017ntire}, Flickr2K \cite{lim2017enhanced}, and LSDIR \cite{li2023lsdir} 
provide synthetic pairs to enable training using downsampled high-resolution images,
while real pairs have also been made available by changing the camera focus on the same scenes \cite{kohler2019toward, cai2019toward}.
However, these datasets do not consider remote sensing images that have a very different overhead view and a much coarser spatial resolution (e.g., meter-level) compared to regular photos taken by hand-held devices.
In the remote sensing domain, the high cost of high-resolution imagery and the challenges of accurately geo-registering corresponding low-resolution data make it difficult to construct large-scale SR datasets  \cite{qi2026advancing}. As a result, most existing studies rely on synthetic datasets generated by applying bicubic downscaling to available remote sensing datasets \cite{qi2026advancing}. More recently, efforts have been made to construct real-world SR datasets by simulating realistic degradations based on the imaging characteristics of spectral sensors  \cite{wang2024towards}. In addition, several remote sensing SR datasets, including PROBA-V \cite{martens2019super}, SEN2VENµS \cite{michel2022sen2venmus}, and WorldStrat \cite{cornebise2022open} have provided paired real low- and high-resolution samples from various sensors (e.g., Landsat-8 and Sentinel-2). 
However, these benchmark datasets all focus on the visual appearance of super-resolved satellite images following benchmarking literature on general non-satellite images. To the best of our knowledge, no existing datasets have integrated the important aspect of downstream tasks in Earth monitoring into the benchmarks, limiting progress in developing SR methods tailored for practical, task-relevant applications. Thus, GeoSR aims to bridge a major gap in the current benchmark datasets for SR tasks using remote sensing images.

\section{GeoSR: From Fidelity to Downstream Task Performance}

The objective of this paper is to go beyond super-resolution in remote sensing and the typical fidelity metrics by integrating a variety of downstream tasks from different application domains to enable direct evaluation of the utility of enhanced image sharpness in the application context.
To achieve this, we will first introduce a new benchmark dataset, GeoSR-Bench, that connects the SR tasks with downstream tasks to support the evaluation, and then discuss the evaluation protocols for developing performance benchmarks and answering important questions between traditional fidelity metrics and downstream task performance.

\begin{figure}[t]
  \centering
  \includegraphics[width=0.48\textwidth]{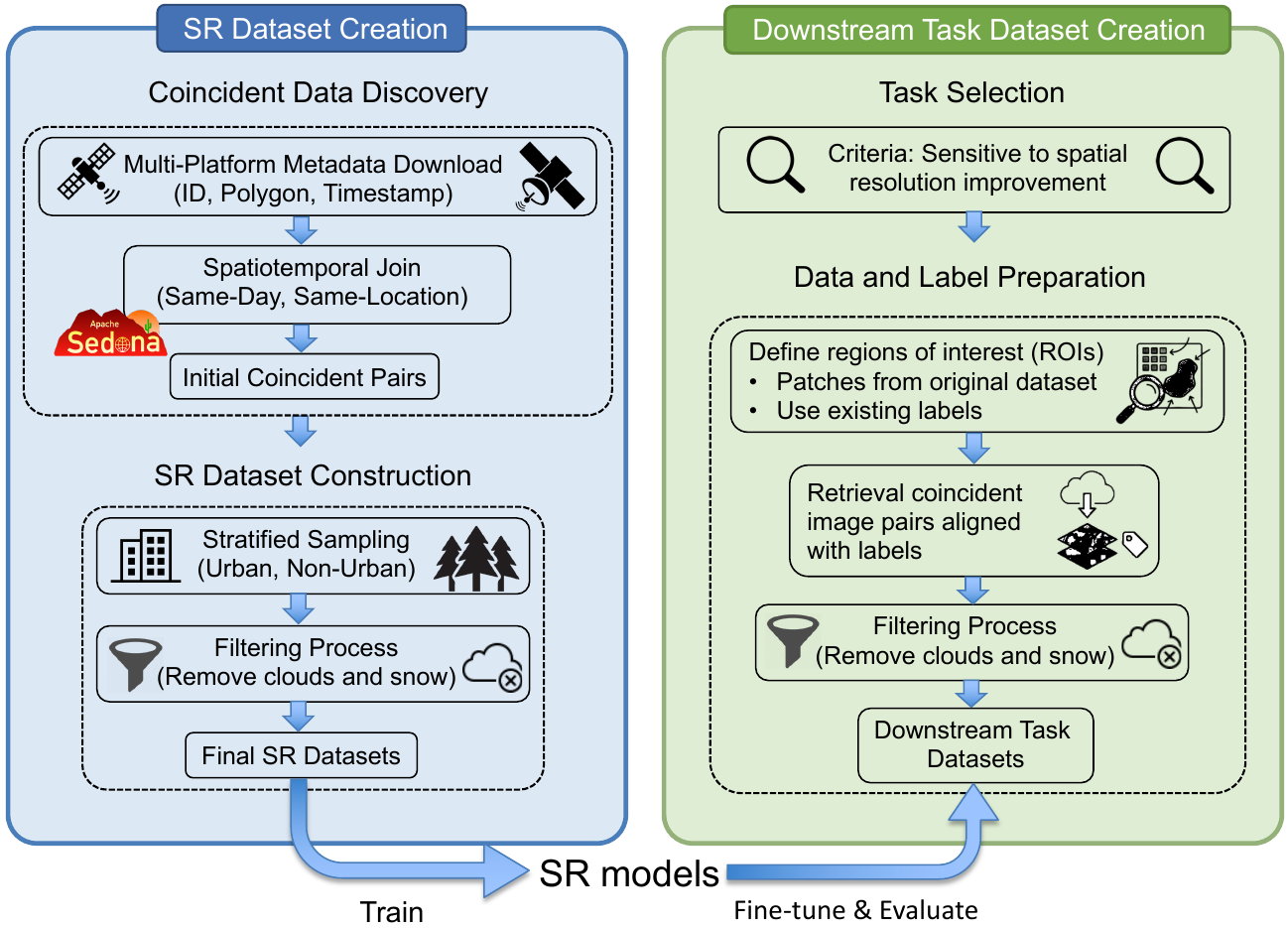}
  \caption{GeoSR Dataset Construction Process Overview.}
  \label{fig:geosr_overview}
\end{figure}

More specifically, the GeoSR-Bench dataset consists of two main components as shown in Fig. \ref{fig:geosr_overview}: (1) The SR component that supports the training of SR models;
and (2) The downstream task component that supports further finetuning and evaluation using a diverse set of downstream applications.
In the following, Sec. \ref{sec:sr} and Sec. \ref{sec:dt} introduce the design of the SR and downstream task components, respectively; Sec. \ref{sec:process} describes additional details for data processing and quality control; and Sec. \ref{sec:eval} discusses the evaluation protocols and main questions.

\subsection{Super-resolution tasks and dataset creation}\label{sec:sr}

To ensure broad resolution coverage, GeoSR-Bench includes multiple widely-used and publicly available remote sensing platforms with coarse, medium, and high resolution. 
Table \ref{tab:platform} lists the platforms that we consider here: coarse-resolution MODIS, medium resolution Landsat-8/Sentinel-2, and high-resolution NAIP.
Based on these, we define two cross-platform SR tasks: 
(1) Coarse-to-medium: From MODIS (500m) to Landsat-8 (30m), and 
(2) Medium-to-high: Sentinel-2 (10m) to NAIP (0.6m).
Note that MODIS itself uses different resolutions across different spectral bands, i.e., 2 bands at 250m resolution, 5 bands at 500m resolution, and others at 1000m resolution. 
We use 500m for MODIS for the first SR task since the set of bands in MODIS at 500m resolution overlap the most with the bands in the target Landsat-8.
Additionally, this also makes the ratio of lower- to higher-resolution images the same across the two SR tasks (i.e., both need to increase the resolution by $\sim$16.7 times). Unlike single-platform SR with synthetic degradation, these cross-platform tasks involve differences in both spatial resolution and sensor characteristics (e.g., spectral response, acquisition geometry, and imaging conditions), which reflects real-world scenarios where SR models must account for more realistic degradations to reconstruct images resembling observations from higher-resolution platforms.

Overall, the two SR tasks collectively cover common SR tasks in the resolution range from 500m to 0.6m as shown in Fig. \ref{fig:sr_settings}.
These two SR tasks also cover diverse utilities. For example, the coarse-to-medium resolution SR task can help capture large-scale vegetation phenology, water monitoring, or land cover transitions, whereas 
the medium-to-high resolution SR task can facilitate the detection of small-scale features such as roads, buildings, and detailed vegetation structures, where the precision of spatial details directly impacts downstream task performance.

\begin{table*}[t]
\centering
\caption{Summary of satellite and aerial imagery platforms used in this study.}
\scriptsize
\begin{tabular}{p{1.5cm} p{1.5cm} p{1.5cm} p{2cm} p{2cm} p{7cm}}
\hline
Platform & Resolution & No. of Bands & Revisit Frequency & Coverage & Product \\
\hline
MODIS & 500\,m & 7 & Daily & Global & MOD09GA.061 Terra Surface Reflectance  \\
Landsat-8 & 30\,m & 7 & 16 days & Global & Landsat 8 Level 2, Collection 2, Tier 1 Surface Reflectance \\
Sentinel-2 & 10\,m & 12 & 5 days & Global & Sentinel-2 Level-2A Surface Reflectance  \\
NAIP & 0.6\,m & 4 & 1–3 years (state-dependent) & Conterminous US & NAIP Product on GEE (USDA/NAIP/DOQQ) \\
\hline
\end{tabular}
\label{tab:platform}
\end{table*}

There are two important considerations when building the datasets for the SR tasks. 
First, the pairs of images collected from the lower- and high-resolution platforms need to be both spatially and temporally aligned to allow the SR models to learn the matched patterns at higher resolution. While the spatial alignment is easy to understand as the pairs of images need to correspond to the same geographic domain, the temporal element is also important.
In particular, 
different remote sensing platforms often operate on different acquisition schedules and revisit frequencies. 
If the images are too far apart, changes over time caused by reflectance difference (e.g., wetter conditions due to rainfall), vegetation dynamics, and so on may generate inconsistency in the image pairs and cause confusion in SR models.
Second, the images need to be representative for training the SR models. 
In the SR context, it means the data should cover areas where SR is commonly needed instead of regions (e.g., tropical forests or large open water) with no clear distinction between lower- and higher- resolution images.
The following sections introduce the details of a coincident data query process and land cover-based sampling to ensure the desired properties of the SR data.

\subsubsection{Spatially and temporally aligned coincident image pairs}
Coincident pairs are defined as two images captured by different satellite platforms that (1) cover the same geographic location and (2) are acquired within a specified temporal window (e.g., within one day). Given data from platforms with different revisit intervals (e.g., daily for MODIS, 5 days for Sentinel-2, 16 days for Landsat-8, and 1–3 years for NAIP), such coincident pairs can be discovered through spatio-temporal joins across tens of millions of images. 
Specifically, we use Apache Sedona \cite{apache_sedona_2025}, an open-source distributed spatial-temporal computing system,
to efficiently identify all coincident pairs at scale. This spatio-temporal join process does not require the full image data, which may take PBs of storage; instead, it operates solely on lightweight metadata such as image IDs, spatial footprints, and capture timestamps. 
We retrieve the Sentinel-2, Landsat-8, and MODIS metadata from the Google Earth Engine (GEE) API, and NAIP metadata from the seamline files provided by USDA \cite{naip_seamlines} that define image boundaries and capture times. These metadata streams serve as inputs to our pipeline for discovering coincident image pairs across the two SR settings, as illustrated in Fig. \ref{fig:geosr_overview}.

We further filter the coincident pairs 
by discarding those with minimal intersection areas (e.g., less than $60\times60$ pixels for the lower-resolution platform or $1000\times1000$ pixels for the higher-resolution platform).
Since the original remote sensing image tiles are often very large (e.g., 10,000 $\times$ 10,000), 
we subdivide the intersection region of each coincident pair into non‑overlapping patches to extract images for SR training and evaluation.
The choice of the patch size is a balanced consideration of both lower- and higher-resolution platforms. Given the roughly 16$\times$ resolution difference, a large size patch for the lower resolution image could make the corresponding patch size too large for the higher resolution image (e.g., from $300\times300$ to $5000\times5000$). On the other hand, the lower resolution image cannot be too small to avoid losing spatial context information.
Considering these factors, we use $60\times60$ as the patch size for the lower-resolution platforms in each SR task, and $1000\times1000$ as the size for the higher-resolution platforms. For MODIS to Landsat-8, each paired patch covers \(30{,}000\,\mathrm{m}\times30{,}000\,\mathrm{m}\), with temporal coverage spanning 2018–2024. For Sentinel-2 to NAIP, each paired patch covers \(600\,\mathrm{m}\times600\,\mathrm{m}\), with temporal coverage spanning 2018–2023.

\subsubsection{Sampling based on land cover for more complete representation}

Here we aim to make sure the SR datasets cover important land cover types or scenarios where SR is commonly needed. 
Since the acquisition locations of the sensing platforms are often uniformly distributed across space (e.g., continental US or global coverage depending on the platform), random sampling would 
make vast majority of samples in remote areas where SR is less needed, such as forest, grassland, open water, etc. These homogeneous landscapes often contain relatively lower-level of structural details and may not help SR models learn the most useful SR skills. In contrast, urban areas often only account for a very small proportion of the overall land, but they contain high-level of detailed textures where resolution enhancement tends to be more needed and effective,
such as building segmentation, impervious surface mapping, and road extraction. 
Therefore, ensuring adequate samples of urban scenes is essential for training super-resolution models that are effective on these resolution-sensitive downstream tasks. 

To account for these, we apply stratified sampling based on existing land cover reference maps, 
and in both SR tasks, we sample urban and non-urban patches using equal proportions.
For the coarse-to-medium SR task (MODIS-to-Sentinel-2),
the dataset has a global coverage as both datasets are publicly available at the global scale.
Here the urban areas are identified using Meta’s Global Urban Areas product \cite{meta_global_urban_areas}, and any patch that either fully overlaps or has the majority of its area intersecting with urban regions in this product is classified as an urban patch. We have also explored the use of ESA WorldCover build-up classes to filter urban areas but found it leads to limited urban patches. This is because the land cover classification scheme used in ESA WorldCover excludes vegetated urban areas from the built-up mask, and we cannot simply merge the built-up class with tree cover or grassland classes as it can introduce confusion with other non-urban forest/grass-covered regions. For completeness and accuracy, we use the Meta’s Global Urban Areas product that is more focused on the urban areas that include all urban land covers.
For non-urban patches, we performed uniform sampling across a diverse set of land cover types defined by the ESA WorldCover classes \cite{esa_worldcover}. 

For the medium-to-high SR task (Sentinel-2-to-NAIP), the dataset has a US coverage, as the high-resolution USDA NAIP imagery by design provides coverage over the US.
Here we use National Land Cover Database (NLCD) \cite{nlcd2021} as the base land cover map because NLCD is a publicly available land cover product focused on the US. Unlike global products (e.g., ESA WorldCover, Meta’s Global Urban Areas), NLCD provides finer-grained classification types such as developed open space, low/medium/high intensity developed areas, primary/secondary/tertiary/thinned road, and deciduous/evergreen forest. These detailed classes are useful for the sampling of diverse urban/non-urban types, which can appear in the medium-to-high downstream tasks that require finer-grained labels.
For urban regions, a patch is classified as urban if at least 50\% of its pixels are labeled as “developed” in NLCD. Urban patches are then evenly sampled across development intensity classes (e.g., open space, low, medium, and high intensity) and key impervious surface categories (e.g., roads and energy infrastructure), which are associated with high-frequency textures and sharp edges. For non-urban regions, we evenly sample from a diverse set of land cover types, including forest, shrubland, grassland, pasture, wetlands, etc.

\begin{figure*}[t]
  \centering
  \includegraphics[width=0.7\textwidth]{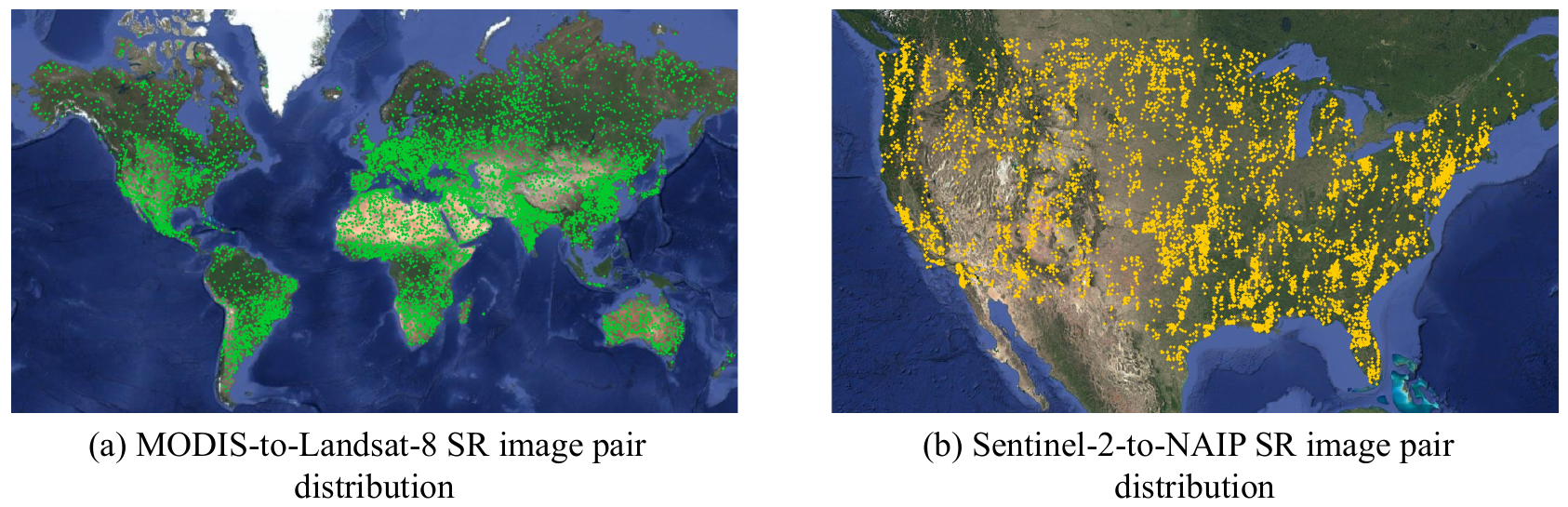}
  \caption{Spatial distributions of paired images in the coincident SR datasets. Green points in (a) indicate MODIS-to-Landsat-8 image-pair locations worldwide, and yellow points in (b) indicate Sentinel-2-to-NAIP image-pair locations across the conterminous US.}
  \label{fig:geosr_distribution}
\end{figure*}

Finally, we sample about 10,000 
pairs of lower- and higher-resolution images for each of the two SR tasks, resulting in about 20,000 pairs in total and collectively covering over 9.3 million $km^2$ of diverse geographic regions. 
The spatial distributions of the image pairs 
for the two SR tasks
are shown in Fig. \ref{fig:geosr_distribution}.
All the images included in the dataset have passed through additional quality control steps that will be summarized later in Sec. \ref{sec:process}.

\subsection{Downstream tasks and dataset creation}\label{sec:dt}

The goal of GeoSR-Bench is to enable evaluation and comparison of SR models in the context of real-world downstream applications. To achieve this, we integrate a diverse set of downstream tasks into the benchmark dataset.
One important selection criterion for a downstream task is that the increase in additional spatial granularity must bring the potential to significantly improve the solution quality for the task. In other words, images from the higher resolution platform must have a clear advantage over images from the lower resolution platform to allow meaningful comparisons between SR models and demonstrate SR benefits.

\begin{table*}[t]
\centering
\caption{Downstream task datasets for the MODIS to Landsat-8 setting in GeoSR-Bench. Each dataset contains labels aligned with coincident image pairs from a lower-resolution platform (MODIS) and a higher-resolution platform (Landsat-8).}
\label{tab:geosr_downstream_m2l}
\begin{adjustbox}{width=1\textwidth}
\scriptsize
\begin{tabular}{p{2.1cm}p{2.1cm}p{2.1cm}p{2.1cm}p{2.2cm}p{4cm}p{0.5cm}}
\hline
Dataset & Task Type & Label Resolution & Label Source  & Location & No. of Samples & Year\\
\hline
River & Semantic segmentation & 30 m  & Dynamic Surface Water Extent (DSWE) \cite{usgs_dswe} & WWF HydroSHEDS \cite{wwf_ffr_gee} & 1065 samples (852 train / 213 test) & 2020 \\
Urban & Semantic segmentation & 10 m  & ESA WorldCover \cite{esa_worldcover} & Global Urban Areas \cite{meta_global_urban_areas} & 1687 samples (1350 train / 337 test) & 2021 \\
Cropland Data Layers (CDL) & Semantic segmentation & 30 m  & Cropland Data Layer \cite{usda_nass_cdl} & US cultivated crop regions & 762 samples (610 train / 152 test) & 2021 \\
Gross Primary Production (GPP) & Regression (GPP estimation) & 30 m  & Landsat GPP CONUS \cite{gee_ntsg_landsat_gpp} & Continental US & 755 samples (604 train / 151 test) & 2021 \\
CHM-S2 & Regression (canopy height) & 10 m  & Global canopy height model \cite{lang2023high} & Global tree areas defined by ESA WorldCover \cite{esa_worldcover} & 1408 samples (1126 train / 282 test)& 2020 \\
\hline
\end{tabular}
\end{adjustbox}
\end{table*}

\begin{table*}[ht]
\centering
\caption{Downstream task datasets for the Sentinel-2 to NAIP setting in GeoSR-Bench. Each dataset contains labels aligned with coincident image pairs from a lower-resolution platform (e.g., Sentinel-2) and a higher-resolution platform (e.g., NAIP).}
\label{tab:geosr_downstream}
\begin{adjustbox}{width=1\textwidth}
\scriptsize
\begin{tabular}
{p{2.1cm}p{2.1cm}p{2.1cm}p{2.1cm}p{2.2cm}p{4cm}p{0.5cm}}
\hline
Dataset & Task Type & Label Resolution & Label Source & Location & No. of Samples & Year \\
\hline
USBuildings & Semantic segmentation & Vector (rasterized at 1 m) & Microsoft Building Footprints \cite{team2018computer} & Southern Wisconsin & 2000 patches (1600 train / 400 test) & 2018 \\
Road Detection & Semantic segmentation & Vector (rasterized with width attributes) & Microsoft US Roads \cite{us_road_detections} & Southern Texas  & 2000 patches (1600 train / 400 test) & 2020 \\
Vermont LC & Semantic segmentation & 0.5 m  & Vermont Land Cover \cite{vermont_land_cover} & Vermont & 2000 patches (1600 train / 400 test) & 2021 \\
ChesapeakeRSC & Semantic segmentation & 1 m  & Chesapeake benchmark \cite{robinson2024seeingroadstreesbenchmark} & Chesapeake Bay region & 2000 patches (1600 train / 400 test) & 2018 \\
CHM-NAIP & Regression (canopy height) & 1 m  & CHM-NAIP dataset \cite{allred2025canopy} & California & 2000 patches (1600 train / 400 test) & 2018-2021 \\
\hline
\end{tabular}
\end{adjustbox}
\end{table*}

Based on this, we select five downstream tasks for each of the two SR tasks, i.e., coarse-to-medium and medium-to-high, as summarized in Tables \ref{tab:geosr_downstream_m2l} and \ref{tab:geosr_downstream}.
As discussed in the scope at the end of Sec. \ref{sec:intro}, we focus on pixel-level tasks, which are widely used in Earth observation applications, including land cover mapping, environmental monitoring, and biophysical variable estimation. To maintain a focused and consistent evaluation framework, we consider both classification and regression tasks, and leave other important applications, such as object detection, instance segmentation, change detection, disaster monitoring, and temporal analysis, for future extensions of the benchmark.
For example, the \textit{Road Detection} dataset in Table \ref{tab:geosr_downstream} is for pixel-level classification, where a high spatial resolution is important to clearly see the boundaries of the roads and generate the correct segmentation masks (e.g., country roads or neighborhood roads). On the other hand, the \textit{CHM-NAIP} dataset is for pixel-level regression to predict the canopy height values of geospatial objects, where high-resolution texture is important to determine the fine-granularity variation of heights in both man-made and natural objects (e.g., buildings, trees).
The downstream tasks in Table \ref{tab:geosr_downstream_m2l} capture events that can be effectively represented at 30m resolution, such as the boundaries of urban areas, river systems, and crop fields. On the other hand, the tasks in Table \ref{tab:geosr_downstream} require much higher resolution details to capture the fine-granularity details of geospatial events and objects such as buildings and road boundaries, as well as local vegetation dynamics.
These tasks span diverse geographic regions and cover a variety of different downstream applications.
In general, classification labels are more commonly available compared to regression, as pixel-level labels with continuous changes over an entire study area are more challenging to generate than discrete class labels, we include two regression tasks in Table \ref{tab:geosr_downstream_m2l} and one in Table \ref{tab:geosr_downstream} based on data available and appropriate for our evaluation, as well as the coverage of different use cases.
In addition, in real-world Earth monitoring applications, the resolution of labels can vary especially at high resolution, depending on the labor and sensing cost (e.g., high-resolution aerial LiDAR). Thus, for the downstream tasks for coarse-to-high resolution, several tasks have labels at 1m resolution instead of the exact original NAIP image resolution at 0.6m. For these tasks, we evaluate the model performance at the original label resolution. Considering the resolution gap from 10m to 0.6m between Sentinel-2 and NAIP images, the 1m resolution is still quite effective in demonstrating the resolution impact.

For the labels (e.g., roads, buildings, croplands), we find and collect them from existing open datasets.
Most datasets come with both labels and images (from the higher-resolution platforms in the two SR tasks), and we use the same coincident data search process described in Sec. \ref{sec:sr} to identify images from the lower-resolution platforms (i.e., MODIS and Sentinel-2 in the two SR tasks, respectively) that are closest to the acquisition date of the image from the higher-resolution platform to form the pairs of images needed for the SR part. 
For datasets that only share the labels but not the images, we use the spatial and temporal information from the metadata to find the lower-resolution images, and the same coincident data search approach to further obtain the higher-resolution images to form the pairs.
In all cases, we verify 
the consistency between labels and images 
to ensure accurate downstream evaluation.
These pairs of images for the SR task in each downstream task will be used for finetuning the SR models and evaluating the quality of super-resolved images. 
For example, both images in a pair are needed to compare the performance of super-resolved images with original higher-resolution images on each downstream task, while they can also be used to evaluate the fidelity-based metrics. Sec. \ref{sec:eval} lists all the major evaluation questions and comparisons we will carry out to better understand the model performances in the context of downstream tasks.

For each downstream task, we sample 2,000 triplets of the higher-resolution image, lower-resolution image, and pixel-level label image (e.g., $1000\times1000$ in size) to maintain the ease-of-use of the overall GeoSR-Bench with 10 downstream tasks. For datasets with fewer than 2,000 images in total, we use all of the available data. Overall, we have about 16,000 triplets for the downstream task component.

\subsection{Data processing and quality control}\label{sec:process}

When constructing both the SR and downstream task datasets, ensuring spatially and temporally aligned coincident image pairs is necessary but not sufficient, because the spectral information captured by different platforms can still vary significantly due to differences in acquisition conditions, creating additional difficulty for SR. For example, in the Sentinel-2-to-NAIP SR task, Sentinel-2 imagery is acquired by satellites observing through the atmosphere, making it subject to atmospheric scattering, absorption, and illumination effects. In contrast, NAIP imagery is collected by aerial platforms flying below most of the atmosphere and cloud cover, and is therefore largely free from atmospheric interference. 
In the MODIS-to-Landsat-8 SR setting, although both platforms are satellite-based, their observations are still affected by atmospheric distortions, which can vary depending on viewing conditions such as solar and sensor angles. These viewing geometries typically differ across platforms, even when images are spatially and temporally coincident, complicating spectral alignment.
To improve cross-platform spectral comparability and reduce atmospheric effects that can vary spatially and temporally, we use surface reflectance products whenever available. For satellite-based sensors such as MODIS, Landsat-8, and Sentinel-2, we utilize atmospherically corrected surface reflectance products provided in GEE.
Although NAIP does not provide a formal surface reflectance product in GEE, its low-altitude, cloud-free acquisition conditions result in relatively stable spectral characteristics, making it suitable as a high-resolution target for training SR models.
Additionally, we retain all available spectral bands of the selected product for each platform as all of the bands can contain auxiliary information to reconstruct spatial details with certain types (e.g., shortwave infrared or red-edge bands can help the SR model better reconstruct land cover types such as vegetation, built-up areas, and water bodies and enhance spatial detail). When different bands from a platform have different native resolutions, we upsample them to the highest available resolution in the product using bilinear interpolation to retain the maximum possible spatial detail, which is commonly done in practice. For example, Sentinel-2’s 20m and 60 m bands are upsampled to the highest 10m resolution to match its RGB and NIR bands; similar data processing is performed for MODIS and Landsat-8 products. For NAIP, as their native resolution varies from 1–2m in older collections to 0.6m or higher in more recent years (e.g., after 2021 for most regions), we use uniformly resampled 0.6m versions available in GEE to ensure spatial consistency across years and regions.
For the temporal consistency, we restrict all datasets (both coincident image pairs for different settings and corresponding labels) to those collected from 2018 onward. This is to ensure that all satellite-based platforms used in this study provide atmospherically corrected surface reflectance (e.g., the availability of Sentinel-2 surface reflectance products begins in March 2017). 

Another source of spectral misalignment arises from different weather conditions, such as clouds and snow, whose dynamic and scene-obscuring nature introduces inconsistencies between coincident image pairs acquired at different times (e.g., 1 day). Moreover, areas covered by extensive clouds or snow often exhibit limited or uniform texture (e.g., scenes entirely covered by clouds or snow), making them less informative for SR training. 
To avoid that, we apply platform-specific cloud and snow filtering using the quality assessment (QA) information provided with each dataset. For satellite-based sensors such as MODIS, Landsat-8, and Sentinel-2, which acquire imagery above the atmosphere, we use the corresponding cloud and snow masks or QA bands (e.g., the MODIS's state\_1km band, the Landsat-8's QA\_PIXEL band, and the Sentinel-2's QA60 band with cloud probability masks \cite{s2_cloud_probability}) to identify and exclude cloudy pixels and retain only cloud-free patches. We retain only image patches with minimal cloud and snow presence by applying strict thresholds (e.g., less than 0.1\%) to balance between data quality and the number of usable image pairs. Notably, a threshold of zero is avoided, as it would be overly restrictive and substantially limit the availability of valid coincident pairs. For NAIP images, as they are collected by aerial platforms flying below cloud layers, they are generally free of cloud contamination. NAIP is also largely free from snow because it’s collected during the leaf-on, snow-free months, with weather-aware flight planning that prioritizes clear, ground-visible conditions. Because of these, no further cloud or snow filtering is applied to NAIP imagery.

After applying cloud and snow filtering as an initial quality control step, we conduct several rounds of visual inspection to improve the dataset quality further. Although the quality assessment (QA) information provided with each product is generally accurate, thin or residual clouds and snow can still remain in some samples. We manually inspect the filtered image pairs to identify any remaining cloud or snow contamination. In addition, we check for significant mismatches between higher- and lower-resolution images caused by temporal changes such as human activities or abrupt natural events (e.g., floods or landslides), which may occur even within short temporal windows.
For datasets where downstream labels are available but the corresponding images are not directly paired, we rely on spatial and temporal metadata to identify coincident lower- and higher-resolution imagery. However, misalignments between imagery and labels may still occur given the chance of label errors or temporal changes in the landscape. To address this, we manually inspect these samples and remove any triplets exhibiting misalignment between higher/lower-resolution images and labels.

\subsection{Evaluation protocol and questions}\label{sec:eval}

GeoSR-Bench enables the evaluation of SR models in the context of downstream applications. This section describes the high-level protocol of the evaluation set-ups and main evaluation questions we aim to answer via the new benchmark dataset.

\subsubsection{Evaluation set-ups}
Given the two main components of the GeoSR-Bench dataset, i.e., the SR task component and the downstream task component, we use a two-stage evaluation that first trains only the SR models using the SR datasets, and then finetunes and tests the SR models' performance on downstream tasks, in combination with a set of corresponding downstream task models.

First, for the SR tasks, we include a diverse set of methods, covering direct transformer-based approaches, GAN-based frameworks, neural operator models, and diffusion models, with the full list of the 9 models in the experiment section (Sec. \ref{sec:exp}).
We train the SR models separately for the two different SR tasks, given the largely different resolution ranges they represent and target. This also aligns well with the practical scenarios, as there exists a large volume of images for each of these remote sensing platforms and each platform often uses their own dedicated models for product generation. 
For each of the two SR tasks (i.e., coarse-to-medium and medium-to-high), we use all samples in the SR datasets to train the models. This stage serves as a pre-training step that produces SR models with strong generalization across diverse land-cover types and imaging conditions (e.g., solar zenith/azimuth angles, seasonal phenology), enabling better adaptation to downstream tasks.

Second, for integration with the downstream tasks, we split the model preparation process into three steps:
\begin{itemize}[leftmargin=*]
    \item SR model finetuning: Since the distribution of images over different types of land cover and geographic regions depends on the nature of each downstream task (e.g., urban vs rural areas), we further finetune the SR models for each of the downstream tasks, respectively, using 80\% of higher/lower-resolution image pairs in the downstream task data for finetuning (with 10\% of the finetuning data reserved for validation) and the remaining 20\% for evaluation. 
    \item Downstream task model training: The SR models by design focus on enhancing image resolution instead of the downstream tasks. Thus, downstream task based models are needed to connect the SR images to the downstream results, and we use 3 widely-used models for pixel-level tasks to generate the final outputs for the downstream tasks: U-Net, SegFormer, and Swin Transformer. Using 3 different models here allows comparisons between the SR models for downstream task performance while helping understand the effects from the downstream task models. 
    These downstream task models are trained using the higher-resolution images with the corresponding labels, as an initial training step.
    \item Downstream task model finetuning: 
    Since the generated SR images from the SR models will be different from the true higher-resolution images, the initial downstream task models trained for each downstream task need to be further finetuned to adapt to the generated images for each of the SR models. This fine-tuning process allows the downstream models to maximize the utility of the generated SR images for downstream tasks while mitigating potential distribution shifts introduced by different SR models.
    For each individual dataset, we use the labels from the same training set to pair with the generated SR images for finetuning.
    In total, we have 9 SR models, 5 downstream tasks, and 3 downstream task models for each of the 2 SR tasks. Given all the training and finetuning, this leads to 270 different model paths.
\end{itemize}

\subsubsection{Evaluation questions}\label{sec:eval_q}

\begin{table}[]
\centering
\caption{Evaluation setups enabled by GeoSR-Bench and their corresponding research questions.}
\label{tab:results}
\scriptsize
\begin{tabular}
{p{0.25\linewidth} p{0.65\linewidth}}
\toprule
Evaluation Setup & Goal and Related Research Questions\\
\midrule
SR results on downstream datasets & 
\textbf{Goal}: Evaluate SR image quality under task-specific domains. \newline
\textbf{Questions}: SR-Q1, SR-Q2, EM-Q1, EM-Q2 \\
\addlinespace
Downstream results on LR images & 
\textbf{Goal}: Serve as the \textit{lower bound} of task performance. \newline
\textbf{Questions}: DT-Q2 \\
\addlinespace
Downstream results on HR images & 
\textbf{Goal}: Serve as the \textit{upper bound} of task performance. \newline
\textbf{Questions}: DT-Q2  \\
\addlinespace
Downstream results on SR images & 
\textbf{Goal}: Evaluate the effectiveness of SR images on downstream tasks. \newline
\textbf{Questions}: DT-Q1, DT-Q2, EM-Q1, EM-Q2 \\
\bottomrule
\end{tabular}
\end{table}

Using all the models from the evaluation set-ups, we can obtain multiple sets of results for different comparison purposes as summarized in Table \ref{tab:results}, where all results are test results on corresponding datasets:
(1) SR results on the downstream task datasets;
(2) Downstream task results using original lower-resolution images;
(3) Downstream task results using original higher-resolution images;
(4) Downstream task results using the generated higher-resolution images.

These results allow us to answer important questions to understand the performance of the SR models on the downstream tasks and how well the standard fidelity metrics can reflect the downstream task utility. 
Specifically, here we list the overarching questions we aim to answer via GeoSR-Bench:
\begin{itemize}[leftmargin=*]
    \item \textbf{SR question 1 (SR-Q1)}: 
    How do different types of SR models perform on remote sensing images in terms of visual and spatial fidelity?
    \item \textbf{SR-Q2}: 
    How well do different types of models generate authentic details for different land cover types?
    \item \textbf{Downstream task question 1 (DT-Q1)}: 
    How effective are the super-resolved images for the downstream tasks and how do the best-performing models change?
    \item \textbf{DT-Q2}: Can the super-resolved images improve downstream task performance over original lower-resolution images (expected lower bound), and how close can they get to the performance of original higher-resolution images (expected upper bound)?
    \item \textbf{Evaluation metric question 1 (EM-Q1)}: How reflective are the visual and spatial fidelity metrics on SR models' performance on downstream tasks? In other words, how well do they correlate in terms of metric values and rankings?
    \item \textbf{EM-Q2}: Under which scenarios are the rankings from fidelity metrics effective or ineffective in indicating rankings based on downstream task utility?
\end{itemize}

\section{Experiments}\label{sec:exp}

To address the evaluation questions outlined in Sec. \ref{sec:eval} using GeoSR-Bench, we conduct a series of experiments including SR performance evaluation, downstream task evaluation, and metric analysis. In the following, Sec. \ref{sec:models} introduces the SR models and downstream task models used throughout the experiments. 
Sec. \ref{sec:fidelity} evaluates SR performance using two types of fidelity metrics: (1) visual fidelity metrics, with PSNR and SSIM; and (2) spatial fidelity metrics, with semivariogram difference.
Sec. \ref{sec:downstream} presents the downstream task results and compares the performance of models trained on SR images against those trained on the original lower- and higher-resolution images. Finally, Sec. \ref{sec:eval_metric} evaluates the relationship between fidelity metrics and the downstream task performance. 

\subsection{Candidate Methods} \label{sec:models}

The following lists the candidate models included for SR and downstream tasks, respectively.

\subsubsection{Super-Resolution}

We select representative SR models across four different categories, where several of the models may include features spanning over categories: (1) Transformer-based models: Models that use a transformer-based architecture to extract global context information for SR; (2) Neural operator-based models: Models treat images as samples from continuous functions and learn mappings between function spaces, rather than fixed-resolution grids; (3) GAN-based models: Models that optimize adversarial losses to produce high-resolution images that are perceptually realistic; and (4) Diffusion-based models: Models generate high-resolution images through a denoising diffusion process, which iteratively refines random noise into detailed outputs conditioned on low-resolution inputs. The list of candidate models includes:
\vspace{-4pt}
\begin{itemize}[leftmargin=*]
\itemsep0em
        \item \textbf{ATD} \cite{zhang2024transcending}: Transformer-based architecture with an Adaptive Token Dictionary (ATD) module to address the limited receptive field of window-based self-attention. 
        \item \textbf{RGT} \cite{chen2023recursive}: Recursive Generalization Transformer (RGT) that recursively aggregates local input features into a global representations that benefit the high-resolution image reconstruction.
        \item \textbf{CFAT} \cite{ray2024cfat}: Transformer-based architecture with a non-overlapping triangular window-based local attention to mitigate boundary-level distortion of the conventional rectangular window approaches.
        \item \textbf{CAMixer} \cite{wang2024camixersr}: A SR architecture with content-aware routing that can dynamically assign simple areas to convolution and complex areas to self-attention.
        \item\textbf{SRNO} \cite{wei2023super}: A model that learns to map low- and high-resolution images as continuous functions, enabling resolution-agnostic super-resolution.
        \item \textbf{ESRGAN} \cite{wang2018esrgan}: Enhanced Super-Resolution Generative Adversarial Networks with an improved perceptual loss for better texture recovery.
        \item \textbf{SeD} \cite{li2024sed}: A GAN-based model with a semantic-aware discriminator that incorporates the semantics from pretrained vision models as the condition. 
        \item \textbf{BiDiff} \cite{chen2024binarized}: A binarized diffusion SR model that reduces the memory and computational overhead of standard diffusion models while preserving performance.
        \item \textbf{UPSR} \cite{zhang2025uncertainty}: A diffusion model that adjusts the noise level across spatial regions based on estimated uncertainty, applying lower noise to smooth areas with low uncertainty.
\end{itemize}

\subsubsection{Downstream Tasks}
To evaluate the effectiveness of super-resolved images on downstream tasks, we consider the following models, which are widely used and have shown good performance in many real-world pixel-level tasks:
\begin{itemize}[leftmargin=*]
\item \textbf{UNet:} An encoder–decoder CNN architecture with skip connections, known for its effectiveness in dense pixel-wise prediction tasks and widespread use in remote sensing applications.
\item \textbf{SegFormer (SegF)} \cite{xie2021segformer}: A transformer-based model that combines a hierarchical encoder with lightweight decoders, offering strong segmentation performance while being computationally efficient.
\item \textbf{Swin Transformer (Swin)} \cite{liu2021swin}: A transformer-based model that captures localized representations by restricting self-attention to shifted local windows for efficient feature extraction. We attach a segmentation head by concatenating multi-scale features from all stages, followed by additional convolutional layers for prediction.
\end{itemize}

\subsection{Evaluation on Super-Resolution Quality} \label{sec:fidelity}

Following the evaluation questions from SR to downstream tasks (Sec. \ref{sec:eval_q}), we start by evaluating SR image quality. Specifically, we compare the SR outputs with their corresponding high-resolution references 
using two types of fidelity metrics: visual fidelity and spatial fidelity. For visual fidelity, we use PSNR to quantify pixel-level accuracy and SSIM to measure local similarity in luminance, contrast, and structure. For spatial fidelity, we use semivariogram differences to measure how spatial dependency changes with distance between locations and evaluate discrepancies from SR images.
Table \ref{tab:fidelity_metrics} summarizes the evaluation focus of each metric. In the following, we present the traditional visual fidelity results (PSNR and SSIM) in Sec. \ref{sec:visual}, followed by the spatial fidelity results in Sec. \ref{sec:spatial}.

\begin{table}[t]
\centering
\caption{Summary of SR image fidelity metrics and their evaluation purposes.}
\label{tab:fidelity_metrics}
\begin{adjustbox}{width=0.48\textwidth}
\scriptsize
\begin{tabular}{ccc}
\hline
Metric & Fidelity type & Evaluation focus \\
\hline
PSNR & Visual fidelity & Pixel-level spectral accuracy \\
SSIM & Visual fidelity & Local luminance, contrast, and structural similarity \\
Semivariogram Diff. & Spatial fidelity & Spatial autocorrelation across scales \\
\hline
\end{tabular}
\end{adjustbox}
\end{table}

\subsubsection{Visual Fidelity Evaluation Using PSNR and SSIM}\label{sec:visual}

\begin{table*}[t]
\centering
\caption{PSNR (dB) on downstream datasets across the two SR tasks.}
\label{tab: psnr}
\begin{adjustbox}{width=1\textwidth}
\begin{tabular}{|c|c||ccccccccc|}
\hline
SR Tasks & Datasets & ATD & RGT & CFAT & CAMixer & SRNO & BiDiff & UPSR & SeD & ESRGAN \\
\hline \hline
\multirow{5}{*}{MODIS to Landsat-8} 
& River        & \textbf{32.83} & 32.80 & 29.98 & 32.61 & 32.68 & 8.47 & 31.07 & 31.80 & 28.70 \\
& Urban        & \textbf{30.76} & 30.68 & 28.19 & 30.54 & 30.67 & 8.81 & 29.51 & 29.68 & 27.00 \\
& CDL          & \textbf{30.01} & 29.94 & 28.00 & 29.91 & 29.95 & 10.01 & 28.79 & 28.43 & 26.62 \\
& GPP          & \textbf{31.13} & 31.02 & 28.99 & 30.96 & 31.05 & 8.76 & 29.82 & 30.05 & 27.39 \\
& CHM-S2       & 35.38 & \textbf{35.45} & 33.19 & 35.19 & 35.34 & 5.48 & 33.85 & 34.01 & 31.53 \\ 
\hline \hline
\multirow{5}{*}{Sentinel-2 to NAIP} 
& USBuildings & 19.07 & \textbf{19.15} & 19.00 & 19.06 & 18.94 & 14.85 & 18.15 & 18.79 & 16.21 \\
& Road Detection & 17.89 & \textbf{18.28} & 17.76 & 17.87 & 17.77 & 10.33 & 17.44 & 17.92 & 15.65 \\
& VermontLC & 18.98 & \textbf{19.24} & 18.83 & 18.84 & 18.76 & 15.19 & 18.30 & 18.40 & 16.41 \\
& ChesapeakeRSC & 17.88 & \textbf{18.05} & 17.90 & 17.85 & 17.84 & 13.54 & 17.34 & 17.82 & 15.48 \\
& CHM-NAIP        & 29.00 & 29.15 & 29.16 & 28.62 & \textbf{29.21} & 17.77 & 25.26  & 26.69 & 23.78 \\
\hline
\end{tabular}
\end{adjustbox}
\end{table*}

\begin{table*}[t]
\centering
\caption{SSIM on downstream datasets across the two SR tasks.}
\label{tab: ssim}
\begin{adjustbox}{width=1\textwidth}
\begin{tabular}{|c|c||ccccccccc|}
\hline
SR Tasks & Datasets & ATD & RGT & CFAT & CAMixer & SRNO & BiDiff  & UPSR & SeD & ESRGAN \\
\hline \hline
\multirow{5}{*}{MODIS to Landsat-8} 
& River        & \textbf{0.77} & \textbf{0.77} & 0.73 & \textbf{0.77} & \textbf{0.77} & 0.14 & 0.74 & 0.75 & 0.63 \\
& Urban        & \textbf{0.71} & \textbf{0.71} & 0.68 & \textbf{0.71} & \textbf{0.71} & 0.11 & 0.68 & 0.69 & 0.59 \\
& CDL          & \textbf{0.73} & \textbf{0.73} & 0.71 & \textbf{0.73} & \textbf{0.73} & 0.15 & 0.70 & 0.70 & 0.59 \\
& GPP          & \textbf{0.75} & \textbf{0.75} & 0.72 & 0.74 & \textbf{0.75} & 0.12 & 0.72 & 0.71 & 0.63 \\
& CHM-S2       & \textbf{0.82} & \textbf{0.82} & 0.79 & 0.81 & \textbf{0.82} & 0.07 & 0.79 & 0.79 & 0.74 \\
\hline \hline
\multirow{5}{*}{Sentinel-2 to NAIP} 
& USBuildings & 0.44 & \textbf{0.45} & 0.44 & 0.44 & 0.44 & 0.35 & 0.39 & 0.41 & 0.32 \\
& Road Detection & 0.37 & \textbf{0.38} & 0.37 & 0.37 & 0.37 & 0.29 & 0.33 & 0.34 & 0.26 \\
& VermontLC & 0.32 & \textbf{0.33} & \textbf{0.33} & \textbf{0.33} & 0.32 & 0.28 & 0.30 & 0.27 & 0.22 \\
& ChesapeakeRSC & \textbf{0.24} & \textbf{0.24} & 0.23 & \textbf{0.24} & 0.23 & 0.18 & 0.21 & 0.22 & 0.15 \\
& CHM-NAIP        & \textbf{0.72} & \textbf{0.72} & \textbf{0.72} & \textbf{0.72} & \textbf{0.72} & 0.61 & 0.58 & 0.68 & 0.60 \\
\hline
\end{tabular}
\end{adjustbox}
\end{table*}

Tables \ref{tab: psnr} and \ref{tab: ssim} demonstrate the PSNR and SSIM results across the evaluated SR models on testing datasets of the downstream tasks. Overall, the MODIS-to-Landsat-8 downstream tasks show higher PSNR and SSIM values than the Sentinel-2-to-NAIP tasks, likely because Landsat-8 imagery is spatially less detailed and contains fewer high-frequency structures than NAIP as the training target, making it easier to reconstruct.
Among SR models, the transformer-based (e.g., ATD and RGT) and neural operator-based models (e.g., SRNO) show higher PSNR and SSIM, whereas GAN-based models (e.g., SeD and ESRGAN) yield lower scores, and diffusion-based models perform the worst under these metrics. This can be attributed to the different training objectives of these SR models: the transformer-based and neural operator-based models rely on pixel-wise L1 loss, which directly minimizes the average absolute difference between the predicted and ground-truth pixels, thus favoring pixel-wise difference-based metrics such as PSNR. On the other hand, GANs and diffusion models optimize objectives that emphasize perceptual realism or generative diversity rather than strict pixel-level accuracy, leading to outputs that are visually plausible but may exhibit larger spectral or texture differences from the ground truth.

\begin{figure}[t]
  \centering
  \includegraphics[width=0.48\textwidth]{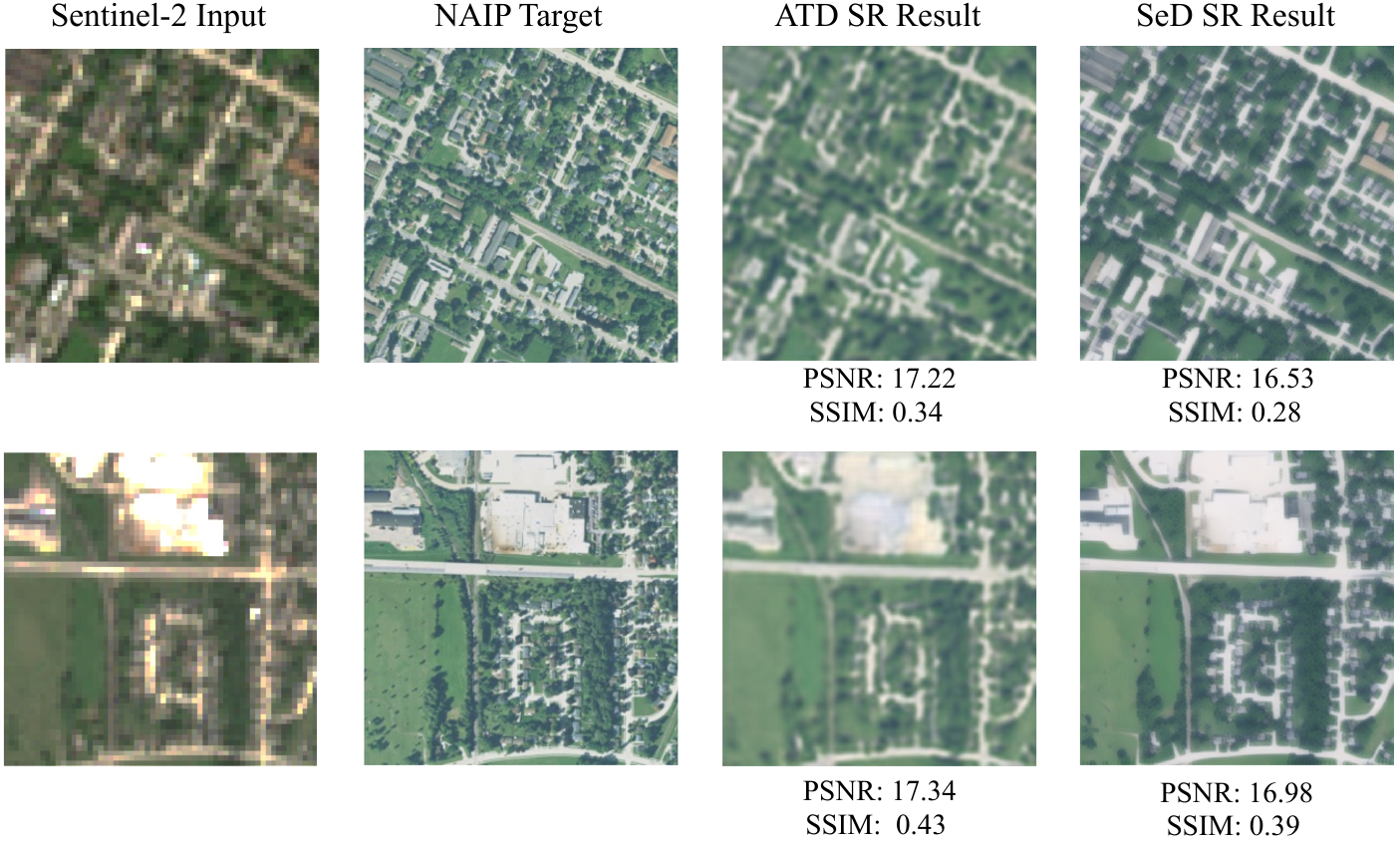}
  \caption{PSNR and SSIM do not always align with visual perception in SR. The results are from the test set of the USBuildings dataset.}
  \label{fig:psnr_ssim_issue}
\end{figure}

By comparing SR performance across datasets, we can evaluate how well SR models reconstruct details for different land covers that each dataset focuses on. In the MODIS-to–Landsat-8 task, the tree/forest-focused \textit{CHM-S2} dataset achieves the highest PSNR and SSIM,
suggesting that forested areas are relatively spectrally and texturally homogeneous, making them easier to reconstruct. The \textit{River} dataset follows, likely because of its relative spectral homogeneity of water surfaces. For other datasets, such as \textit{Urban}, \textit{CDL}, and \textit{GPP}, they contain more complex and heterogeneous land covers, making them harder to reconstruct and resulting in lower PSNR and SSIM values. The \textit{Urban} dataset includes complex textures and sharp edges associated with man-made structures, and cropland in \textit{CDL} consists of fine-grained, spatially heterogeneous field patterns. The \textit{GPP} dataset includes mixed land cover types, making it more challenging than datasets dominated by a single dominant class, such as \textit{River} or \textit{CHM-S2}. 
A similar trend is observed in the Sentinel-2–to-NAIP task. The tree-focused \textit{CHM-NAIP} dataset achieves the highest PSNR and SSIM, followed by datasets dominated by single major land cover types, such as \textit{USBuildings} and \textit{Road Detection}. Datasets with highly mixed land covers, such as \textit{VermontLC} and \textit{ChesapeakeRSC}, are more difficult to reconstruct and exhibit relatively lower PSNR and SSIM values.

It is important to note that, despite their widespread adoption in SR evaluation, PSNR and SSIM are known to correlate imperfectly with human perception. Prior work \cite{zhang2018unreasonable, wolters2023zooming, su2025rethinking} shows that both metrics can favor over-smoothed or blurry reconstructions that align with pixel values or local structural statistics, but may not be visually desirable in SR tasks. As illustrated in Fig. \ref{fig:psnr_ssim_issue}, the SeD model generates more perceptually detailed outputs, but receives lower PSNR and SSIM scores compared to the more blurry outputs from the ATD model.  
Given these limitations, our dataset enables the evaluation of SR models not only in terms of visual fidelity but also through their downstream task performance, providing a more application-oriented assessment of SR effectiveness.

\subsubsection{Spatial Fidelity Evaluation Using Semivariogram Difference}\label{sec:spatial}

Our spatial fidelity evaluation is based on semivariograms, which is a standard measure for spatial dependency as a function of distance between locations.
Following the spatial evaluation protocol in \cite{qiu2023evaluation}, we randomly sample 1\% of pixels from each image to reduce computational cost. For each sampled pixel, we pair it with pixels located at lag distance $h$ in four directions (upper, lower, left, and right). The semivariogram at lag distance $h$ is calculated as
\begin{equation}
\gamma(h)=\frac{1}{2N(h)}
\sum_{(i,j)\in P(h)}
\left[z(\mathbf{s}_i)-z(\mathbf{s}_j)\right]^2,
\end{equation}
where $z(\mathbf{s}_i)$ and $z(\mathbf{s}_j)$ are pixel values at locations
$\mathbf{s}_i$ and $\mathbf{s}_j$, $P(h)$ is the set of valid pixel pairs separated by distance $h$ in the four directions, and $N(h)$ is the number of such pairs. A semivariogram typically increases with lag distance as spatial dependency weakens, eventually reaching a plateau where spatial dependence becomes negligible. Its shape therefore characterizes the strength and spatial extent of dependency across different distances.

We then define the semivariogram difference as the absolute difference between the semivariograms of the SR predictions and their corresponding high-resolution reference images at each spatial lag distance: $|\Delta\gamma(h)| =
|\gamma_{\mathrm{SR}}(h)-\gamma_{\mathrm{HR}}(h)|$.
The differences are averaged across spectral bands and evaluation samples. Before the plateau, these differences indicate how well SR preserves spatial autocorrelation, while beyond the plateau, they primarily capture discrepancies in the plateau values. Thus, smaller differences indicate higher fidelity in preserving the spatial dependency structure of the reference imagery.
To summarize the differences across all lag distances, we calculate
the area under the semivariogram-difference curve (AUC), where a
smaller AUC indicates higher spatial fidelity. We rank the SR methods by
their AUC within each dataset and report the average rank across the five
datasets.

\begin{figure*}[t]
    \centering
    \includegraphics[width=\textwidth]{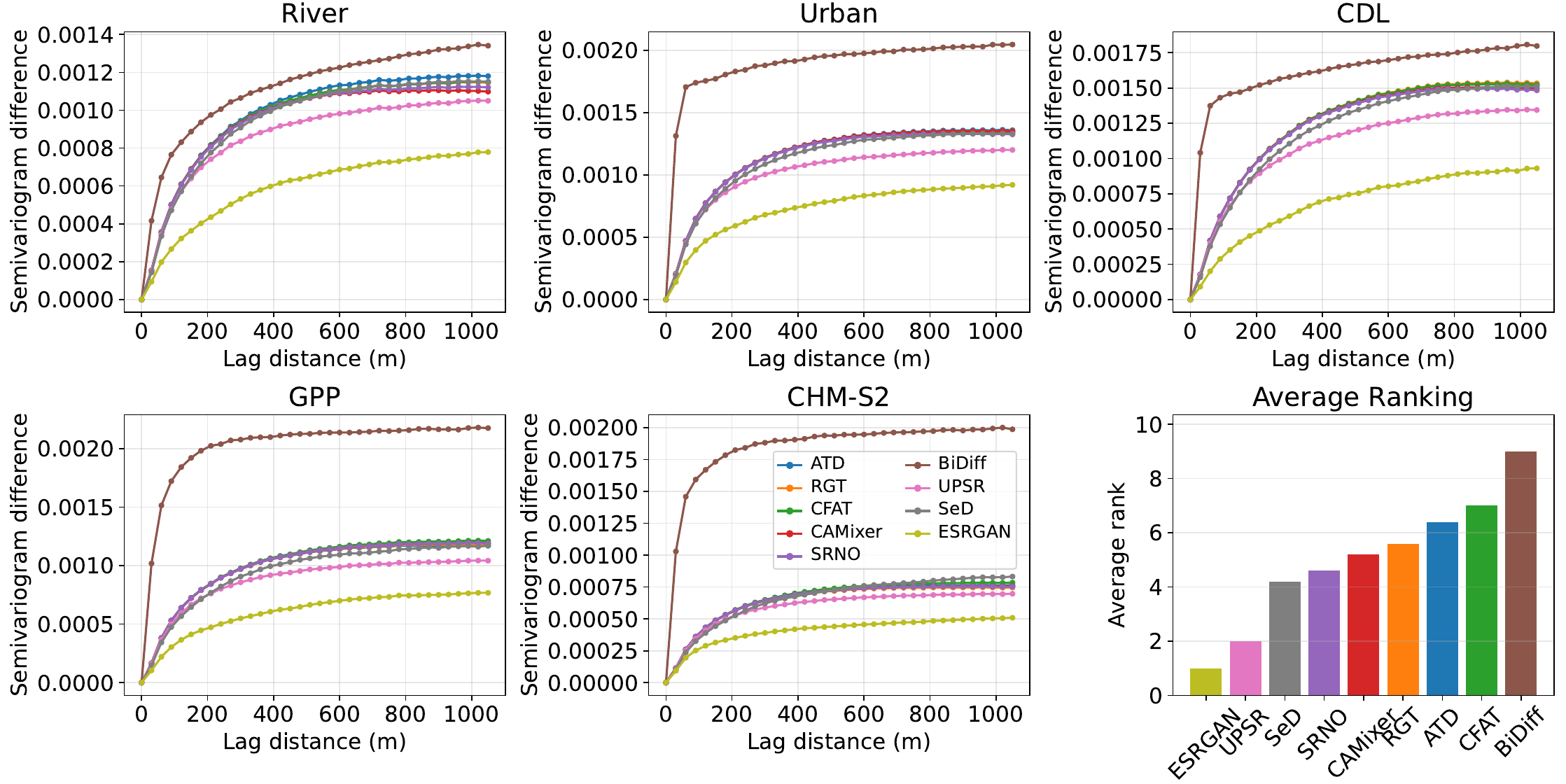}
    \caption{Spatial fidelity of SR methods on the MODIS-to-Landsat-8 task. Each dataset panel shows the absolute semivariogram difference between the SR prediction and the original high-resolution image across lag distances, where lower values indicate better preservation of spatial variability.
    The bottom-right panel reports the average AUC-based rank of each method across the five datasets, with lower ranks indicating better overall spatial fidelity.}
    \label{fig:spatial_fidelity_1}
\end{figure*}

\begin{figure*}[t]
    \centering
    \includegraphics[width=\textwidth]{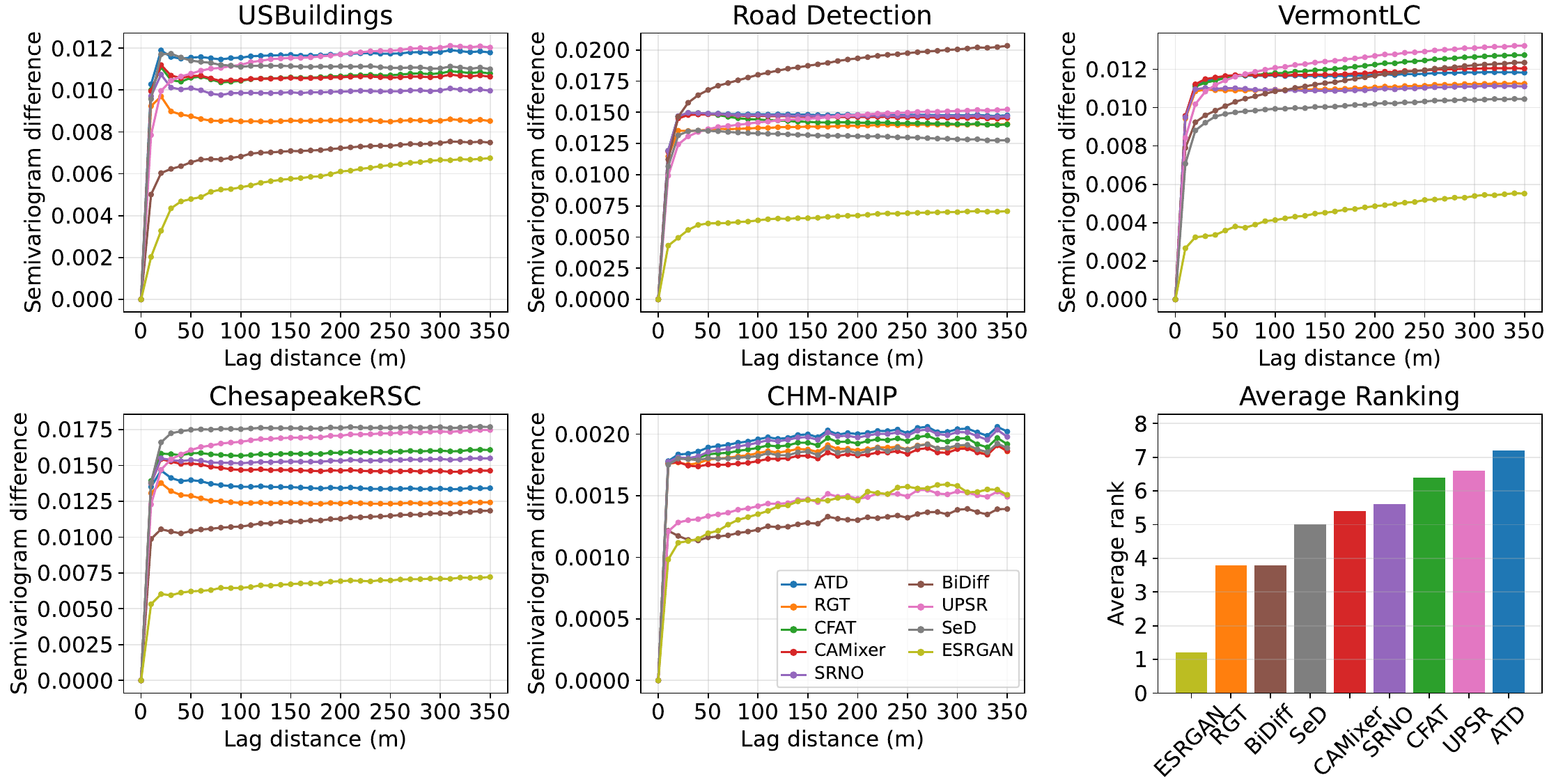}
    \caption{Spatial fidelity of SR methods on the Sentinel-2-to-NAIP task. Each dataset panel shows the absolute semivariogram difference between the SR prediction and the original high-resolution image across lag distances, where lower values indicate better preservation of spatial variability.
    The bottom-right panel reports the average AUC-based rank of each method across the five datasets, with lower ranks indicating better overall spatial fidelity.}
    \label{fig:spatial_fidelity_2}
\end{figure*}

As shown in Figs.~\ref{fig:spatial_fidelity_1} and \ref{fig:spatial_fidelity_2}, the semivariogram differences between the SR predictions and the original high-resolution images vary considerably across the two SR tasks. For MODIS-to-Landsat-8, ESRGAN achieves the smallest semivariogram differences across datasets, followed by UPSR and SeD on average, whereas BiDiff exhibits the largest spatial discrepancies. For Sentinel-2-to-NAIP, ESRGAN remains the best overall, while RGT and BiDiff rank next, and ATD shows the largest discrepancies. The substantial changes in method rankings across the two SR tasks indicate that the ability to recover spatial patterns is highly resolution- and dataset-dependent. Moreover, the spatial results differ substantially from the spectral fidelity results in Table \ref{tab: psnr}: ESRGAN consistently achieves strong spatial fidelity despite relatively poor PSNR, while ATD generally shows the opposite tendency. These contrasting patterns suggest that preserving spectral information does not necessarily imply better recovery of spatial structure, and that the relationship between the two can vary with the SR task.

\subsection{Downstream Task Evaluation} \label{sec:downstream}

\subsubsection{Downstream Task Performance for the MODIS-to-Landsat-8 SR} 

\begin{table*}[]
\caption{Performance of pixel-level prediction models evaluated on MODIS-to-Landsat-8 downstream tasks (averaged over three independent runs). 
For binary classification datasets such as River, we calculate the F1-score of the target class. For multiclass datasets such as Urban LC and  CDL, we calculate the mean F1-score over all classes. For regression datasets GPP and Temperature, we calculate the Mean Absolute Error (MAE) as the evaluation metric.
}
\label{tab:m2l8}
\begin{adjustbox}{width=1\textwidth}
\centering
\begin{tabular}{|c|c||c|ccccccccc|c|}
\hline
Datasets & Predictive & \multirow{2}{*}{MODIS} & \multicolumn{9}{c|}{Super-resolution models} & \multirow{2}{*}{LS8} \\
 (Eval. Metric)  &  Models  &       &     ATD &     RGT &    CFAT & CAMixer &    SRNO &  BiDiff &  UPSR &   SeD &  ESRGAN &    \\
\hline
\hline
\multirow{3}{*}{\shortstack{River\\(F1 $\uparrow$)}} 
& UNet & 0.68 & \srcell{\textbf{0.74}} & \srcell{0.73} & \srcell{0.72} & \srcell{0.73} & \srcell{\textbf{0.74}} & 0.56 & \srcell{0.69} & \srcell{0.72} & 0.67 & 0.95 \\
& SegF & 0.65 & \srcell{0.73} & \srcell{\textbf{0.75}} & \srcell{0.73} & \srcell{0.74} & \srcell{0.74} & \srcell{0.66} & \srcell{0.70} & \srcell{0.72} & \srcell{0.68} & 0.94 \\
& Swin & 0.74 & \textbf{0.74} & \textbf{0.74} & \textbf{0.74} & 0.73 & \textbf{0.74} & 0.61 & 0.70 & 0.72 & 0.67 & 0.94 \\
\hline
\hline
\multirow{3}{*}{\shortstack{Urban LC\\(Mean F1 $\uparrow$)}} 
& UNet & 0.34 & \textbf{0.32} & 0.31 & 0.31 & 0.31 & \textbf{0.32} & 0.23 & 0.29 & 0.28 & 0.26 & 0.44 \\
& SegF & 0.35 & \srcell{\textbf{0.39}} & \srcell{0.38} & \srcell{0.37} & \srcell{0.36} & \srcell{0.38} & 0.28 & 0.32 & 0.33 & 0.29 & 0.50 \\
& Swin & 0.41 & \textbf{0.39} & 0.38 & 0.37 & 0.37 & \textbf{0.39} & 0.27 & 0.33 & 0.35 & 0.29 & 0.49 \\
\hline
\hline
\multirow{3}{*}{\shortstack{CDL\\(Mean F1 $\uparrow$)}} 
& UNet & 0.45 & 0.39 & 0.39 & 0.37 & 0.37 & \textbf{0.41} & 0.16 & 0.25 & 0.16 & 0.22 & 0.57 \\
& SegF & 0.43 & \srcell{\textbf{0.49}} & \srcell{0.45} & \srcell{0.47} & \srcell{0.47} & \srcell{0.48} & 0.21 & 0.33 & 0.30 & 0.18 & 0.65 \\
& Swin & 0.52 & 0.50 & 0.49 & 0.49 & \textbf{0.51} & 0.50 & 0.20 & 0.34 & 0.36 & 0.21 & 0.66 \\
\hline
\hline
\multirow{3}{*}{\shortstack{GPP\\(MAE $\downarrow$, $\times10^{-2}$)}} 
& UNet & 3.26 & \srcell{2.17} & \srcell{2.19} & \srcell{2.31} & \srcell{2.15} & \srcell{\textbf{2.13}} & \srcell{2.84} & \srcell{2.40} & \srcell{2.57} & \srcell{2.48} & 2.89 \\
& SegF & 3.18 & \srcell{2.65} & \srcell{2.64} & \srcell{2.73} & \srcell{\textbf{2.38}} & \srcell{2.45} & \srcell{2.61} & \srcell{2.58} & \srcell{2.68} & 3.35 & 2.85 \\
& Swin & 3.12 & \srcell{2.24} & \srcell{\textbf{2.17}} & \srcell{2.38} & \srcell{2.22} & \srcell{2.20} & \srcell{2.95} & \srcell{2.95} & \srcell{2.33} & \srcell{2.58} & 2.82 \\
\hline
\hline
\multirow{3}{*}{\shortstack{CHM-S2\\(MAE $\downarrow$)}} 
& UNet & 6.74 & \srcell{5.22} & \srcell{5.36} & \srcell{5.86} & \srcell{5.39} & \srcell{\textbf{5.18}} & 8.04 & \srcell{6.03} & \srcell{6.10} & \srcell{6.50} & 5.16 \\
& SegF & 7.14 & \srcell{4.93} & \srcell{\textbf{4.88}} & \srcell{5.38} & \srcell{5.01} & \srcell{4.99} & 7.78 & \srcell{5.49} & \srcell{5.61} & \srcell{6.05} & 5.14 \\
& Swin & 6.41 & \srcell{5.09} & \srcell{\textbf{5.05}} & \srcell{5.54} & \srcell{5.14} & \srcell{5.17} & 7.47 & \srcell{5.56} & \srcell{5.63} & \srcell{6.15} & 4.99 \\
\hline
\multicolumn{12}{l}{Note: SR downstream task performance that surpasses the MODIS baseline is colored in green.}\\
\end{tabular}
\end{adjustbox}
\end{table*}

\begin{figure*}[t]
  \centering
  \includegraphics[width=0.98\textwidth]{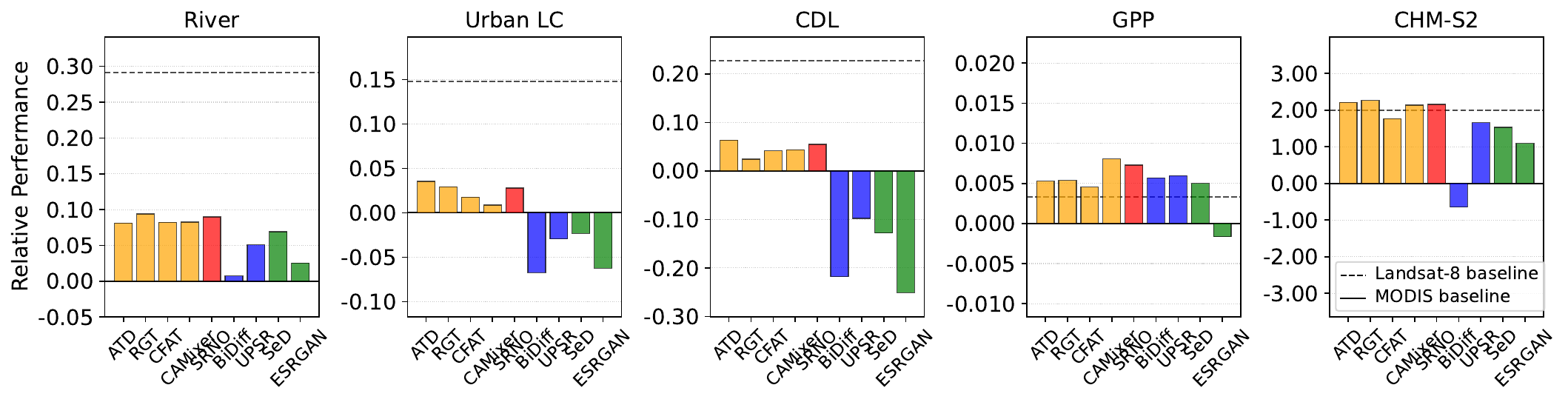}
  \caption{Relative performance comparison across SR models on the MODIS-to-Landsat-8 downstream tasks using SegFormer as the downstream model. Relative Performance from the original MODIS imagery (expected lower bound) is adjusted to 0, and the performance of the original Landsat-8 imagery (expected upper bound) is expressed as the improvement over the lower bound, indicated by the dashed lines. For regression datasets using MAE metric, the MAE values are sign-flipped so that higher values indicate better performance, aligning with the direction of classification metrics. Different groups of SR models are shown in different colors for clarity.}
  \label{fig:relative_m2l8}
\end{figure*}

Table \ref{tab:m2l8} presents the results on the pixel-wise classification and regression downstream tasks under the MODIS-to-Landsat-8 SR task. 
For each downstream dataset, downstream task performance is evaluated and compared across the original MODIS inputs, the generated SR images, and the original Landsat-8 (LS8) images. 
To enhance result robustness, for each evaluation case we train the pixel-level prediction models three times, and report performance averaged over the three runs. 

In Table \ref{tab:m2l8}, the cells with a green background are cases where the SR images improve downstream task performance compared to the baseline lower-resolution images.
As we can see, there are still many cases where the downstream task performance based on super-resolved images is lower than that of the lower resolution images. 
As discussed in Sec. \ref{sec:eval}, the downstream task models were finetuned using the SR images, so the degradation in performance is not simply caused by distribution shifts caused by SR, but more likely by the incorrect SR patterns generated that have lower alignment with labels.
Among the classification tasks, 
the \textit{River} dataset exhibits the greatest number of improvements, along with the largest performance gain of nearly 10\%. For the other two datasets, the largest performance gains from SR models range from 4\% to 6\% in F1 scores (including mean F1 scores).
Among the three downstream models, SegFormer accounts for the largest number of downstream improvements compared to the other two. 
Despite the gains, the performance gaps between SR and original higher resolution images remain large.
Across SR model types, transformer and neural operator-based models lead to most downstream improvements and diffusion-based models generally lag behind, due to their higher spectral and texture variability in the reconstruction.
Among all SR models, ATD achieves the best F1 scores in most scenarios, though its performance over other transformer-based models (e.g., RGT) is small. In contrast, BiDiff and ESRGAN generally had lower downstream task performance, with performance often below the MODIS baseline, due to a higher level of 
mismatching 
details and spectral inaccuracies from both models.

For regression tasks on the \textit{GPP} and \textit{CHM-S2} datasets, the performance improvements are more pervasive than those on the classification tasks.
Most SR models achieve lower prediction errors than the original lower-resolution inputs, with RGT most frequently delivering the best performance. Notably, in rare cases, the best-performing SR models even outperform the high-resolution Landsat-8 baselines.
One possible reason is that SR methods produce cleaner and more consistent outputs than real Landsat-8 imagery, reducing spectral noise that can degrade model performance, while still introducing enough fine-grained structure to improve predictions.

Fig. \ref{fig:relative_m2l8} compares the performance of SR images relative to the performance of original lower-resolution MODIS and higher-resolution Landsat-8 images across the five downstream tasks. We select SegFormer as the downstream model in this figure because it shows the largest performance improvements over the lower-resolution baselines among the three downstream task models. 
Different groups of SR models (i.e., transformer-based, neural-operator-based, diffusion-based, and GAN-based models) are shown in different colors for easier visual separation. We calculate the relative performance of SR using $P_{sr}-P_{lower}$, where $P_{sr}$ is the downstream task performance of an SR model, and $P_{lower}$ is the downstream task performance of the original lower-resolution MODIS images. For reference, we adjust the MODIS baseline to zero and the Landsat-8 baseline to $P_{higher}-P_{lower}$ (shown as dashed lines), where $P_{higher}$ is the performance of the original Landsat-8 images. For regression datasets, the performance values (e.g, MAE) are \textbf{sign-flipped} so that results have the same direction as those from classification tasks.
Overall, a substantial performance gap remains between SR models and the original higher-resolution imagery, particularly for classification datasets. The best-performing models such as ATD and SRNO only recover less than one-third of the relative performance gap between the lower- and higher-resolution imagery in most cases, while BiDiff and ESRGAN often show negative relative performance. This highlights considerable room for improvement in the downstream utility of SR methods.

\subsubsection{Downstream Task Performance for the Sentinel-2-to-NAIP SR}

\begin{table*}[]
\caption{Performance of pixel-level prediction models evaluated on Sentinel-2-to-NAIP downstream tasks (averaged over three independent runs). For binary classification datasets such as USBuildings and Road Detection, we calculate the F1-score of the target class. For multiclass datasets such as Vermont LC and  ChesapeakeRSC, we calculate the mean F1-score over all classes. For the regression dataset CHM-NAIP, we calculate the Mean Absolute Error (MAE) as the evaluation metric.}
\label{tab:f1_s2naip}
\begin{adjustbox}{width=1\textwidth}
\centering
\begin{tabular}{|c|c||c|ccccccccc|c|}
\hline
Datasets & Predictive & \multirow{2}{*}{S2} & \multicolumn{9}{c|}{Super-resolution models} & \multirow{2}{*}{NAIP} \\
 (Eval. Metric)  &  Models  &       &     ATD &     RGT &    CFAT & CAMixer &    SRNO &  BiDiff &  UPSR &   SeD &  ESRGAN &    \\
\hline
\hline
\multirow{3}{*}{\shortstack{USBuildings\\(F1 $\uparrow$)}} 
& UNet & 0.31 & \srcell{0.38} & \srcell{\textbf{0.46}} & \srcell{0.37} & \srcell{0.40} & \srcell{0.37} & 0.27 & \srcell{0.33} & \srcell{0.41} & 0.28 & 0.82 \\
& SegF & 0.12 & \srcell{0.42} & \srcell{\textbf{0.46}} & \srcell{0.43} & \srcell{0.40} & \srcell{0.38} & \srcell{0.32} & \srcell{0.34} & \srcell{0.44} & \srcell{0.29} & 0.82 \\
& Swin & 0.50 & 0.47 & 0.50 & 0.47 & 0.45 & 0.44 & 0.36 & 0.33 & \srcell{\textbf{0.51}} & 0.39 & 0.82 \\
\hline
\hline
\multirow{3}{*}{\shortstack{Road Detection\\(F1 $\uparrow$)}} 
& UNet & 0.57 & 0.50 & \srcell{\textbf{0.62}} & 0.53 & 0.53 & 0.49 & 0.30 & \srcell{0.59} & \srcell{0.61} & 0.48 & 0.86 \\
& SegF & 0.39 & \srcell{0.56} & \srcell{0.63} & \srcell{0.57} & \srcell{0.54} & \srcell{0.54} & \srcell{0.44} & \srcell{0.61} & \srcell{\textbf{0.65}} & \srcell{0.48} & 0.86 \\
& Swin & 0.64 & 0.62 & \srcell{0.68} & 0.62 & 0.61 & 0.59 & 0.50 & 0.64 & \srcell{\textbf{0.69}} & 0.57 & 0.87 \\
\hline
\hline
\multirow{3}{*}{\shortstack{Vermont LC\\(Mean F1 $\uparrow$)}} 
& UNet & 0.35 & 0.33 & 0.35 & 0.32 & 0.34 & 0.32 & 0.26 & \srcell{0.36} & \srcell{\textbf{0.38}} & 0.29 & 0.45 \\
& SegF & 0.34 & \srcell{0.38} & \srcell{0.40} & \srcell{0.38} & \srcell{0.38} & \srcell{0.35} & 0.32 & \srcell{0.38} & \srcell{\textbf{0.41}} & 0.33 & 0.52 \\
& Swin & 0.46 & 0.40 & 0.44 & 0.40 & 0.44 & 0.41 & 0.36 & 0.40 & 0.45 & 0.36 & 0.57 \\
\hline
\hline
\multirow{3}{*}{\shortstack{ChesapeakeRSC\\(Mean F1 $\uparrow$)}} 
& UNet & 0.36 & 0.31 & 0.33 & 0.30 & 0.30 & 0.30 & 0.24 & 0.32 & \textbf{0.34} & 0.30 & 0.43 \\
& SegF & 0.28 & \srcell{0.35} & \srcell{0.35} & \srcell{0.35} & \srcell{0.35} & \srcell{0.35} & \srcell{0.29} & \srcell{0.35} & \srcell{\textbf{0.36}} & \srcell{0.31} & 0.47 \\
& Swin & 0.39 & \srcell{\textbf{0.40}} & \srcell{\textbf{0.40}} & 0.38 & 0.38 & 0.39 & 0.32 & 0.37 & \srcell{\textbf{0.40}} & 0.35 & 0.51 \\
\hline
\hline
\multirow{3}{*}{\shortstack{CHM-NAIP\\(MAE $\downarrow$, $\times10^{-2}$)}} 
& UNet & 1.85 & 2.07 & 2.10 & \textbf{1.91} & 2.03 & 2.15 & 3.01 & 2.17 & 2.07 & 2.01 & 1.55 \\
& SegF & 2.07 & 3.61 & 5.13 & 8.59 & 3.18 & 4.49 & 3.20 & 2.36 & 3.04 & \textbf{2.88} & 1.65 \\
& Swin & 2.47 & \srcell{\textbf{2.27}} & \srcell{2.42} & \srcell{2.30} & 2.48 & 2.99 & \srcell{2.31} & 2.58 & \srcell{2.42} & 2.54 & 1.70 \\
\hline
\multicolumn{12}{l}{Note: SR downstream task performance that surpasses the Sentinel-2 baseline is colored in green.}\\
\end{tabular}
\end{adjustbox}
\end{table*}

\begin{figure*}[t]
  \centering
  \includegraphics[width=1.0\textwidth]{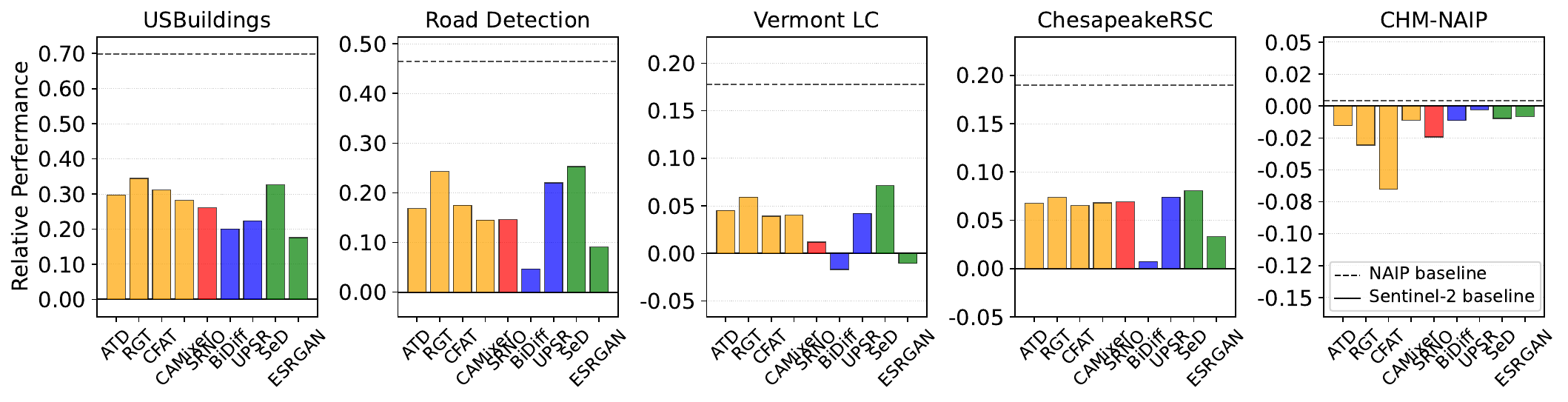}
  \caption{Relative performance comparison across SR models on the Sentinel-2-to-NAIP downstream tasks using SegFormer as the downstream model. Relative performance from the original Sentinel-2 imagery is adjusted to 0 (expected lower bound), and that from the original NAIP imagery (expected higher bound) is represented by the dashed lines as a reference. For regression datasets using MAE metric, the MAE values are sign-flipped so that higher values consistently indicate better performance, aligning with the direction of classification metrics. Different groups of SR models are shown in different colors for clarity.}
  \label{fig:relative_s2naip}
\end{figure*}

Table \ref{tab:f1_s2naip} presents the results on the pixel-wise downstream datasets for the Sentinel-2 to NAIP SR. In this task, the higher-resolution NAIP and SR images have limited four bands but much finer spatial resolution (0.6m) that is smaller than most common objects (e.g., buildings, roads, and trees), making texture information more important for the downstream tasks. 
For classification downstream tasks, compared to the MODIS-to-Landsat-8 setting, there are more cases in which SR-based downstream task performance surpasses the lower-resolution baseline, likely because Sentinel-2 inputs provide richer spatial (i.e., 10m vs. 500m), enabling SR models to more effectively recover spatial structures. 
Among downstream models, SegFormer still accounts for the majority of improvements over the lower-resolution baseline. Across SR models, the GAN-based SeD model achieves the strongest downstream gains in this SR setting, owing to its superior ability to recover fine-scale boundaries and local spatial structures in small geographic features (e.g., buildings and roads, as illustrated in Fig. \ref{fig:psnr_ssim_issue}) that are more important in NAIP-based downstream tasks, with improvements ranging from 4\% to over 30\%.
Despite the improvements over lower-resolution baselines, substantial performance gaps remain compared to the original high-resolution NAIP imagery. These gaps are larger on binary classification tasks (e.g., \textit{USBuilding} and \textit{Road Detection}) than multiclass tasks (e.g., \textit{Vermont LC} and \textit{ChesapeakeRSC}). This may be because multiclass tasks benefit more from globally restored textures and broader spatial context in SR images, whereas binary tasks depend on specific objects occupying only small local regions, which SR models may consider less during the training process.

On the regression dataset \textit{CHM-NAIP}, most SR images do not significantly improve performance over lower-resolution inputs. This is primarily due to the prevalence of fine-scale tree structures (e.g., low shrubs) in this dataset, which are challenging to reconstruct accurately through SR and are essential for precise height prediction. 
Although certain models (e.g., BiDiff) can generate visually detailed SR images that recover small trees, these reconstructed details may not align well with the actual tree locations in the original NAIP imagery or with the ground-truth height labels, resulting in limited performance.

Fig. \ref{fig:relative_s2naip} shows the relative performance from different SR models on the Sentinel-2-to-NAIP downstream tasks. Compared to the MODIS-to-Landsat-8 results in Fig. \ref{fig:relative_m2l8}, SR models in this setting exhibit larger relative improvements over the lower-resolution baselines. In particular, the SeD model recovers over 50\% of the gap on the \textit{Road Detection} dataset. Nevertheless, large gaps remain between SR images and the high-resolution NAIP baselines, underscoring the need for SR models that are designed and optimized with the awareness of downstream tasks.

\begin{figure*}[t]
  \centering
  \includegraphics[width=1\textwidth]{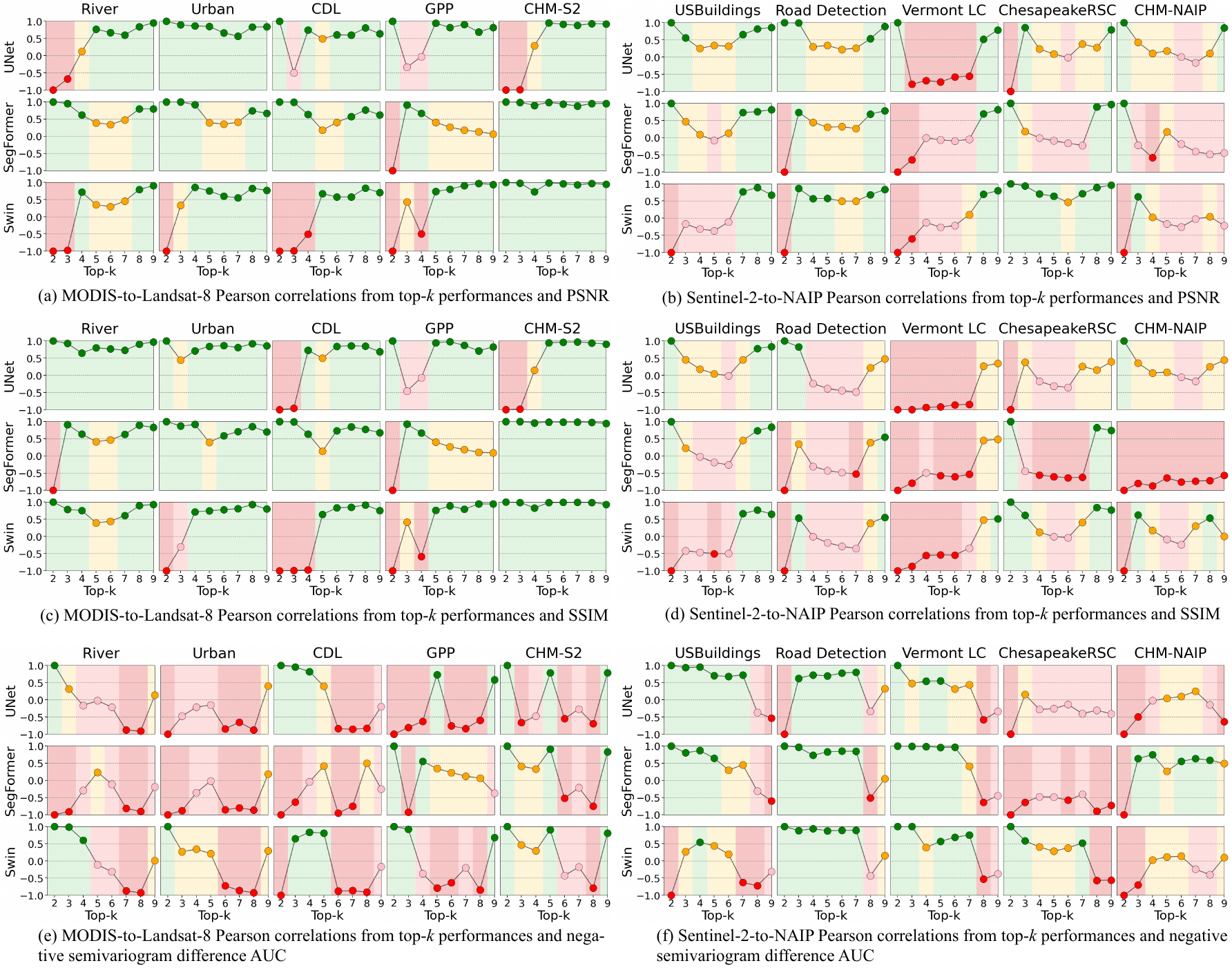}
  \caption{Pearson correlation between fidelity metrics (PSNR/SSIM/semivariogram difference AUC) and downstream task performance computed within the top-$k$ downstream-ranked models. For the ease of comparison, we color-code the correlation values into four categories: strong positive ($r \ge 0.5$, green), weak positive ($0 \le r < 0.5$, yellow), weak negative ($-0.5 \le r < 0$, pink), and strong negative ($ r \le -0.5$, red). }
  \label{fig:p_corr}
\end{figure*}

\begin{figure*}[]
  \centering
  \includegraphics[width=1\textwidth]{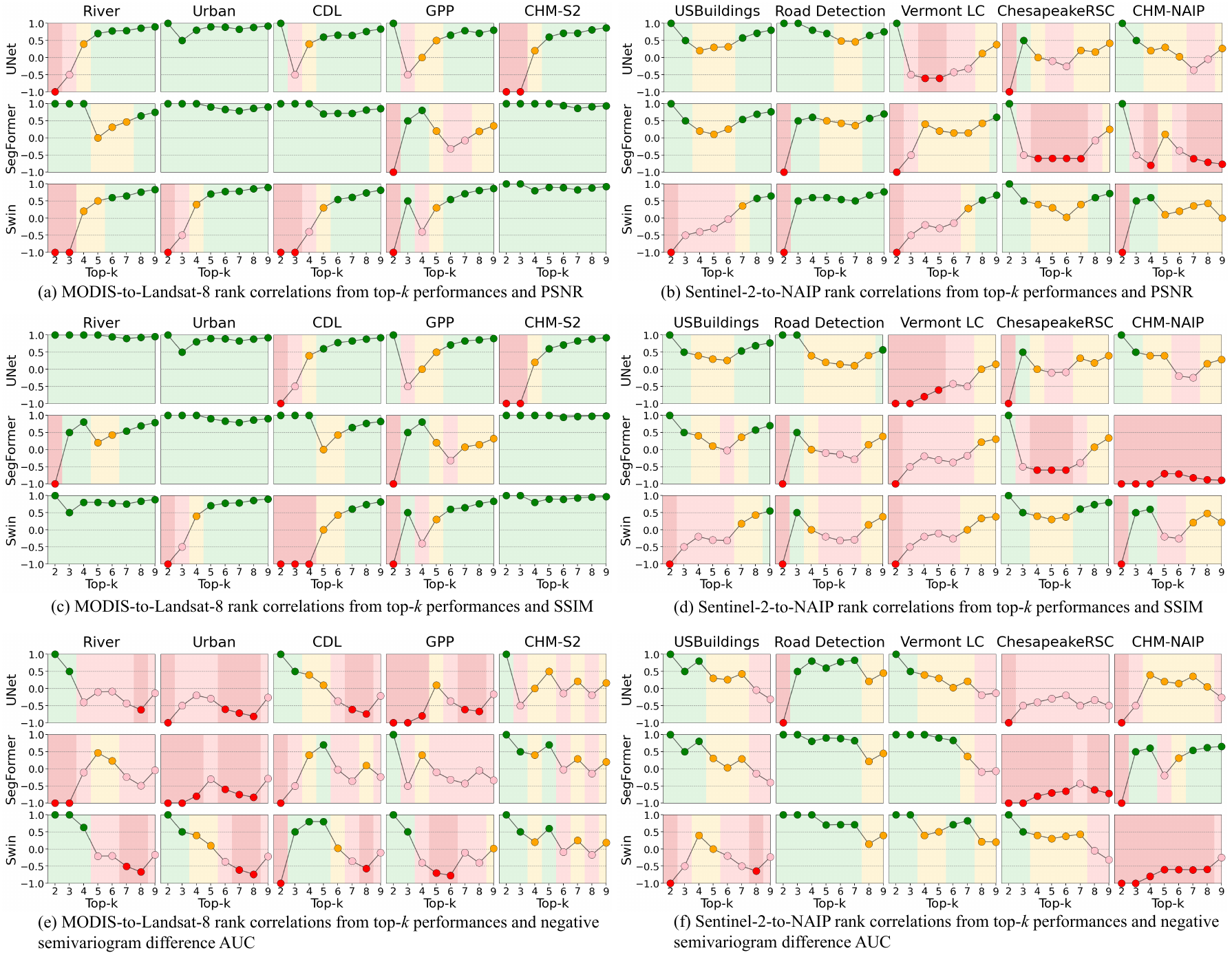}
  \caption{Spearman's rank correlation between fidelity metrics (PSNR/SSIM/semivariogram difference AUC) and downstream task performance computed within the top-$k$ downstream-ranked models. For the ease of comparison, we color-code the rank correlation values into four categories: strong positive ($r \ge 0.5$, green), weak positive ($0 \le r < 0.5$, yellow), weak negative ($-0.5 \le r < 0$, pink), and strong negative ($ r \le -0.5$, red). }
  \label{fig:rank_corr}
\end{figure*}

\subsection{Do SR Fidelity Metrics Imply Downstream Task Performance?} \label{sec:eval_metric}

Given visual and spatial fidelity metrics and downstream task performance across a diverse set of SR models and tasks, one question to be answered is whether the fidelity metrics are reflective on SR models’ performance on downstream tasks, as these fidelity metrics are often used to compare and rank SR models for SR research in both general and remote sensing cases. To address that, we calculate the Pearson correlation between fidelity metrics (PSNR/SSIM/semivariogram difference AUC) and downstream task performance across the 10 benchmark datasets. 
There are several important aspects that we need to consider in the correlation evaluation:
(1) 
Models with performances that are substantially lower than the others tend to dominate the correlation value, as the large distance between a single low-performing model and others' can easily create a global trend, making the differences between other models look insignificant (or noise-like). For example, a very low-performing SR model will likely cause very low downstream task scores as well, creating a low-low anchor point in the fidelity-vs-downstream task scatter plot. If this low-low anchor is far from other models' performance in this plot, a high correlation score is almost guaranteed, regardless of the relationships between the other models.
(2)
When evaluating a model in SR research, the main interest often lies in comparing it with the state-of-the-art models with similar and competitive performances. Very commonly the scrutiny further reduces to the comparison between one model (e.g., newly proposed) and one or two other close runner-ups. This means that, for our analysis, it is important to understand the correlation between a group of competitive, top-ranking methods to understand how well visual- and spatial-fidelity-based rankings align with those from actual downstream tasks.

\textbf{Top-$k$ correlation coefficients:} To further understand under what performance regimes these fidelity metrics align with downstream utility, we introduce a top-$k$ analysis. Specifically, we first rank SR models according to their downstream task performance, and then compute the correlation between downstream task metrics and fidelity metrics within the top-$k$ subset, with the results shown in Fig. \ref{fig:p_corr}.
By progressively expanding $k$, we examine how the correlation evolves as lower-performing models are gradually included.

\textbf{Metric direction alignment:} For classification tasks, higher metric values indicate better performance. In contrast, regression metrics such as MAE follow the opposite direction, where lower values indicate better performance. Similarly, lower semivariogram difference AUC indicates better spatial fidelity. To make the results easily comparable across tasks, we unify the direction of the metrics here by sign-flipping lower-is-better metrics (i.e., $\text{–}$MAE and $\text{–}$semivariogram difference AUC), so that higher values consistently correspond to better performance.

According to Fig. \ref{fig:p_corr},
there similarly exists a large proportion of cases where SR-based metrics do not correlate well with downstream-task-based metrics, with low or even negative correlation scores, especially between smaller competitive groups (i.e., when $k$ is small such as top-3).
Particularly, for visual fidelity (PSNR and SSIM), the correlations generally become stronger and more positive as $k$ increases (e.g., top-8 or top-9), especially for MODIS-to-Landsat-8.  This suggests that the correlation between visual fidelity and downstream performance strengthens as the gap between better-performing and lower-performing SR models expands. In contrast, spatial fidelity, measured by negative semivariogram difference AUC, exhibits more heterogeneous patterns, with generally weak or negative correlations for MODIS-to-Landsat-8 but stronger positive correlations for several Sentinel-2-to-NAIP datasets, suggesting that preserving multi-scale spatial structure may be more relevant to downstream utility for the higher-resolution NAIP tasks.
However, for a group of competitive models that are normally the focus of evaluations in SR model development, the correlations across all these metrics tend to be highly unstable (shown in red, pink, and yellow shaded areas), and the ability of these fidelity metrics to determine downstream task performance is very limited. In those cases, small improvements in fidelity metrics do not necessarily translate into meaningful differences in downstream task performance. Therefore, these metrics provide only limited guidance when concluding superior models for remote sensing downstream tasks.

This trend is further supported by the model ranking analysis shown in Fig. \ref{fig:rank_corr}, which presents the Spearman's rank correlations between downstream-task performance rankings and those based on fidelity metrics. Values closer to 1 indicate stronger agreement between the two ranking criteria.
The visual-fidelity (i.e., PSNR and SSIM) rankings show stronger agreement when lower-performing models are included, whereas the spatial-fidelity rankings remain weakly or negatively correlated with downstream rankings for many datasets, even when lower-performing models are included.
Similarly, the ranking correlations are relatively lower for a group of models with competitive performances (e.g., when $k<5$), indicating frequent mismatches between fidelity-based rankings and downstream task performance rankings. 

\begin{figure}[t]
  \centering
  \includegraphics[width=0.48\textwidth]{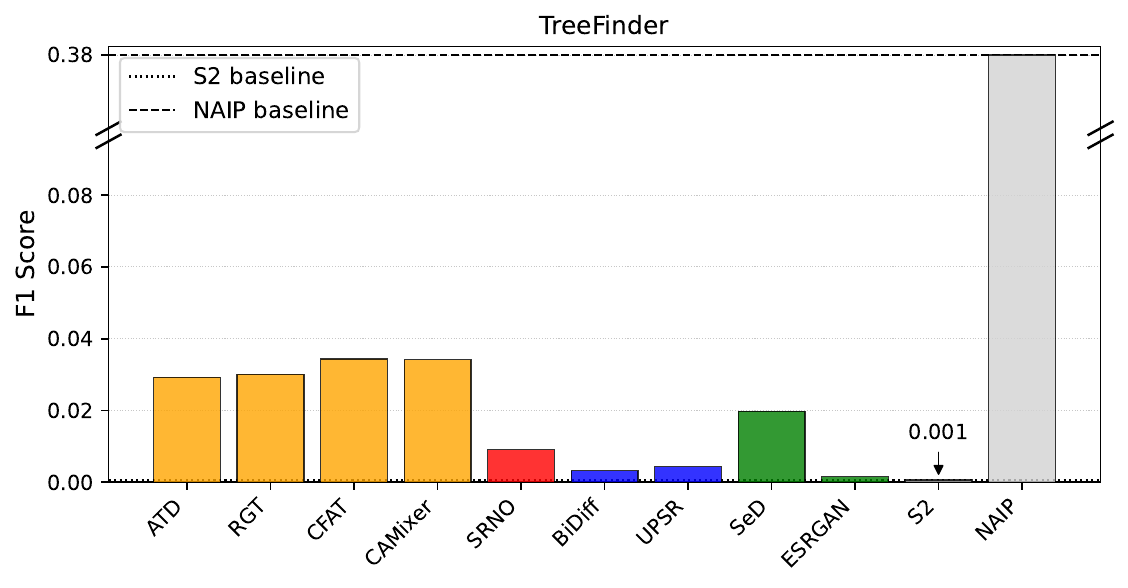}
  \caption{
  Downstream task performance comparison on the TreeFinder dataset using SegFormer as the downstream model. 
  }
  \label{fig:treefinder}
\end{figure}

\section{Discussion: Limit of SR for Sub-Pixel Tasks} \label{sec:challenging_case}


Finally, we discuss the limit or boundary of SR for remote sensing images for sub-pixel events, which is
significantly more challenging  
with targets smaller than a single pixel in the lower-resolution images. In this case, the spectral distinction between the target and its surroundings is significantly reduced due to signal dilution when all the fine-scale signals are aggregated into a coarser pixel. Conceptually, the effectiveness of SR is expected to be minimal for segmentation/mapping tasks where locations are essential. Due to the mixing of signals within a coarse-resolution pixel, it can be infeasible to accurately recover the true locations of the target event given the non-unique solution space.

To provide a concrete example, we carry out an additional evaluation using the TreeFinder dataset \cite{wang2026treefinder}. This dataset aims to map individual tree mortality events (i.e., individual dead trees that may have a crown diameter of 5-10 meters), which are typically larger than NAIP’s 0.6m resolution but smaller than Sentinel-2’s 10m resolution. The original TreeFinder dataset is constructed using NAIP imagery as the input. To align with the settings of other downstream tasks, we sample 2,000 images (1,600 for training and 400 for testing) from this dataset and retrieve the corresponding coincident Sentinel-2 images as the lower-resolution inputs.

Fig. \ref{fig:treefinder} presents the downstream task performance on the TreeFinder dataset. We adopt SegFormer as the downstream model, given its better performance in previous tasks. As shown in the figure, the original Sentinel-2 inputs show very poor performance in detecting fine-scale tree mortality, with performance close to zero. Although SR outputs provide some improvement, the gains are minimal (e.g., 0-4\%), and remain insignificant compared to the performance achieved by original NAIP imagery. 
The results show that for sub-pixel downstream tasks, while SR models may still reach reasonable visual fidelity metrics (e.g., PSNR values of 18.13 for ATD and 18.28 for CFAT, which are comparable to the results in Table \ref{tab: psnr}), the utility for the downstream applications is limited.

\section{Conclusion}

This paper developed GeoSR-Bench, the first benchmark dataset for remote sensing image SR that integrates downstream tasks to evaluate the real-world utility of SR models. GeoSR-Bench provides a large-scale, globally distributed SR dataset composed of temporally aligned, spatially co-located, and quality-controlled image pairs from approximately 36,000 locations worldwide. The dataset covers two cross-platform SR tasks: MODIS to Landsat‑8 (500m to 30m) and Sentinel‑2 to NAIP (10m to 0.6m), which collectively cover spatial resolutions and platforms that are most widely used in downstream applications. Beyond traditional fidelity-based SR evaluation, GeoSR-Bench further incorporates 10 downstream datasets (5 per SR task) with pixel-level labels, allowing SR models to be fine-tuned for downstream tasks and evaluated using task-relevant performance metrics in applications such as 
land cover segmentation, infrastructure mapping, and biophysical variable estimation, bridging a critical gap left by existing SR benchmark datasets. Using GeoSR-Bench, we conducted a comprehensive evaluation of 9 representative SR models, including GAN-, transformer-, neural operator-, and diffusion-based methods, across 270 experimental settings. Our results demonstrate that improvements in traditional fidelity-based SR metrics such as PSNR and SSIM do not always correlate with downstream task performance gains and can even exhibit negative correlations, especially for models that lead to high downstream task performance, where higher fidelity scores do not necessarily imply better task-level utility. These results highlight the limitations of existing benchmarks that rely solely on fidelity metrics to guide SR model development and reveal the necessity of establishing a new evaluation track by emphasizing application-driven performance. 
In the future, we plan to develop new models and metrics to better integrate and consider downstream tasks during the SR process, and expand the benchmarks to cover other types of remote sensing tasks and sensing platforms.

\section*{Acknowledgments}
Zhili Li, Kangyang Chai, Zhihao Wang, and Yiqun Xie are supported in part by the NSF under Grant No. 2126474, 2147195, 2425844, and 2530610; NASA under grant 80NSSC25K0013 and 80NSSC25K7221; Google’s AI for Social Good Impact Scholars program; and the Zaratan cluster at the University of Maryland. Xiaowei Jia is supported in part by the NSF under Grant No. 2239175, 2147195, 2316305, 2425845, 2530609, 2203581; NASA under Grant No. 80NSSC24K1061 and 80NSSC25K0013; the USGS awards G21AC10564 and G22AC00266; and Pitt Momentum Funds and CRC at the University of Pittsburgh. Gengchen Mai is supported by the NSF under Grant No. 2521631. The authors used ChatGPT only for language polishing and grammar improvement.

\bibliographystyle{plain} 
\bibliography{ref}   

\end{document}